\def\cvprPaperID{6866}
\def\confName{CVPR}
\def\confYear{2023}

\def\authorBlock{
    Kim Sung-Bin${}^{1}$ \qquad
    Arda Senocak${}^{2}$ \qquad
    Hyunwoo Ha${}^{1}$ \qquad
    Andrew Owens${}^{3}$ \qquad
    Tae-Hyun Oh${}^{1,4,5}$ \vspace{3mm} \\
    ${}^{1}$Dept.~of Electrical Engineering and ${}^{4}$Grad.~School of Artificial Intelligence, POSTECH\\
    ${}^{2}$Dept.~of Electrical Engineering, KAIST \qquad ${}^{3}$University of Michigan\\ 
${}^{5}$Institute for Convergence Research and Education in Advanced Technology, Yonsei University\\
{\normalsize\url{https://sound2scene.github.io/}}

}

\newif\ifreview 
\newif\ifarxiv \newcommand{\arxiv}{\arxivtrue}
\newif\ifcamera 
\newif\ifrebuttal 
\arxiv 

\pdfoutput=1
\documentclass[10pt,twocolumn,letterpaper]{article}

\ifreview \usepackage[review]{cvpr} \fi
\ifarxiv \usepackage[pagenumbers]{cvpr} \fi
\ifrebuttal \usepackage[rebuttal]{cvpr} \fi
\ifcamera \usepackage{cvpr} \fi

\usepackage{graphicx}
\usepackage{amsmath}
\usepackage{amssymb}
\usepackage{booktabs}

\usepackage{times}
\usepackage{microtype}
\usepackage{epsfig}
\usepackage[table,xcdraw]{xcolor}
\usepackage{caption}
\usepackage{float}
\usepackage{placeins}
\usepackage{color, colortbl}
\usepackage{stfloats}
\usepackage{enumitem}
\usepackage{tabularx}
\usepackage{xstring}
\usepackage{multirow}
\usepackage{xspace}
\usepackage{url}
\usepackage{subcaption}
\usepackage{xcolor}
\usepackage[hang,flushmargin]{footmisc}

\ifcamera \usepackage[accsupp]{axessibility} \fi

% ...

\ifarxiv  \fi

\newcommand{\R}[1]{{%
    \textbf{%
        \ifstrequal{#1}{1}{\textcolor{red}{R#1}}{%
        \ifstrequal{#1}{2}{\textcolor{blue}{R#1}}{%
        \ifstrequal{#1}{3}{\textcolor{magenta}{R#1}}{%
        \ifstrequal{#1}{4}{\textcolor{teal}{R#1}}{%
                           \textcolor{cyan}{R#1}%
        }}}}%
    }%
}}

\usepackage{xr-hyper}

\makeatletter
\newcommand*{\addFileDependency}[1]{
  \typeout{(#1)}
  \@addtofilelist{#1}
  \IfFileExists{#1}{}{\typeout{No file #1.}}
}

\makeatother

\usepackage[pagebackref,breaklinks,colorlinks=True,citecolor=kellygreen,linkcolor=kellygreen,bookmarks=false]{hyperref}

\usepackage[capitalize]{cleveref}
\crefname{section}{Sec.}{Secs.}
\crefname{table}{Table}{Tables}
\crefname{figure}{Fig.}{Figs.}

\usepackage{enumitem}
\usepackage{multirow}
\usepackage{wrapfig}
\usepackage{colortbl}
\usepackage[normalem]{ulem}
\usepackage{comment}
\usepackage{microtype}

\setlist[itemize]{align=parleft,left=0pt,topsep=1mm,itemsep=0mm,parsep=1mm}

\definecolor{azure(colorwheel)}{rgb}{0.0, 0.5, 1.0}
\definecolor{nicegreen}{rgb}{0.0, 0.7, 0.1}
\definecolor{yw}{rgb}{0.01176, 0.5490, 0.5490}
\definecolor{ashblue}{rgb}{0.36, 0.54, 0.66}
\definecolor{ashgrey}{rgb}{0.7, 0.75, 0.71}
\definecolor{applegreen}{rgb}{0.55, 0.71, 0.0}
\definecolor{greenyellow}{rgb}{0.68, 1.0, 0.18}
\definecolor{junebud}{rgb}{0.74, 0.85, 0.34}
\definecolor{kellygreen}{rgb}{0.3, 0.73, 0.09}
\definecolor{ywg}{rgb}{0.9960, 0.8984, 0.5859}
\definecolor{jy}{rgb}{0.58, 0, 0.827}
\definecolor{cornellred}{rgb}{0.7, 0.11, 0.11}
\definecolor{darkcyan}{rgb}{0.0, 0.55, 0.55}
\definecolor{CuGray}{gray}{0.9}
\definecolor{airforceblue}{rgb}{0.36, 0.54, 0.66}
\definecolor{rev}{rgb}{0.784, 0.003, 0.313}
\definecolor{pink}{cmyk}{0, 0.7808, 0.4429, 0.1412}
\definecolor{amethyst}{rgb}{0.6, 0.4, 0.8}
\definecolor{black}{rgb}{0.0, 0.0, 0.0}
\definecolor{tb3_yellow}{rgb}{0.996, 1.0, 0.6}
\definecolor{tb3_orange}{rgb}{0.980, 0.8, 0.604}
\definecolor{tb3_red}{rgb}{0.972, 0.6, 0.6}
\definecolor{dimgray}{rgb}{0.41, 0.41, 0.41}
\definecolor{brickred}{rgb}{0.8, 0.25, 0.33}
\definecolor{bleudefrance}{rgb}{0.19, 0.55, 0.91}
\definecolor{blue(ncs)}{rgb}{0.265, 0.445, 0.765}
\definecolor{green(ncs)}{rgb}{0.0, 0.62, 0.42}

\newcolumntype{g}{>{\columncolor{CuGray}}c}
\newcolumntype{z}{>{\columncolor{CuGray}}l}

\renewcommand{\paragraph}[1]{\vspace{0.5mm}\noindent\textbf{#1.}\,\,}

\usepackage{xspace}

\makeatletter
\def\@fnsymbol#1{\ensuremath{\ifcase#1\or *\or \dagger\or \ddagger\or
   \mathsection\or \mathparagraph\or \|\or **\or \dagger\dagger
   \or \ddagger\ddagger \else\@ctrerr\fi}}
\makeatother

\def\onedot{.\@\xspace}
\def\eg{\emph{e.g}\onedot} 
\def\ie{\emph{i.e}\onedot} 
 
\def\etc{\emph{etc}\onedot} 
 
\def\etal{\emph{et al}\onedot}

\newcommand{\Sref}[1]{Sec.~\ref{#1}}
\newcommand{\Eref}[1]{Eq.~(\ref{#1})}
\newcommand{\Fref}[1]{Fig.~\ref{#1}}
\newcommand{\Tref}[1]{Table~\ref{#1}}

\newcommand{\ba}{{\mathbf{a}}}
\newcommand{\bb}{{\mathbf{b}}}

\newcommand{\be}{\begin{eqnarray}}
\newcommand{\ee}{\end{eqnarray}}
\newcommand{\bee}{\begin{eqnarray*}}
\newcommand{\eee}{\end{eqnarray*}}

\newcommand{\matrixb}{\left[ \begin{array}}
\newcommand{\matrixe}{\end{array} \right]}

\usepackage{mathtools}
\DeclarePairedDelimiter{\norm}{\lVert}{\rVert}
\DeclarePairedDelimiterX{\inp}[2]{\langle}{\rangle}{#1, #2}

\newcommand{\blank}[1]{\hspace*{#1}}

\begin{document}
\title{Sound to Visual Scene Generation by Audio-to-Visual Latent Alignment
}

\author{\authorBlock}
\twocolumn[{
\renewcommand\twocolumn[1][]{#1}%
\maketitle
\centering
    \centering
    \captionsetup{type=figure}
    \includegraphics[width=1\textwidth]{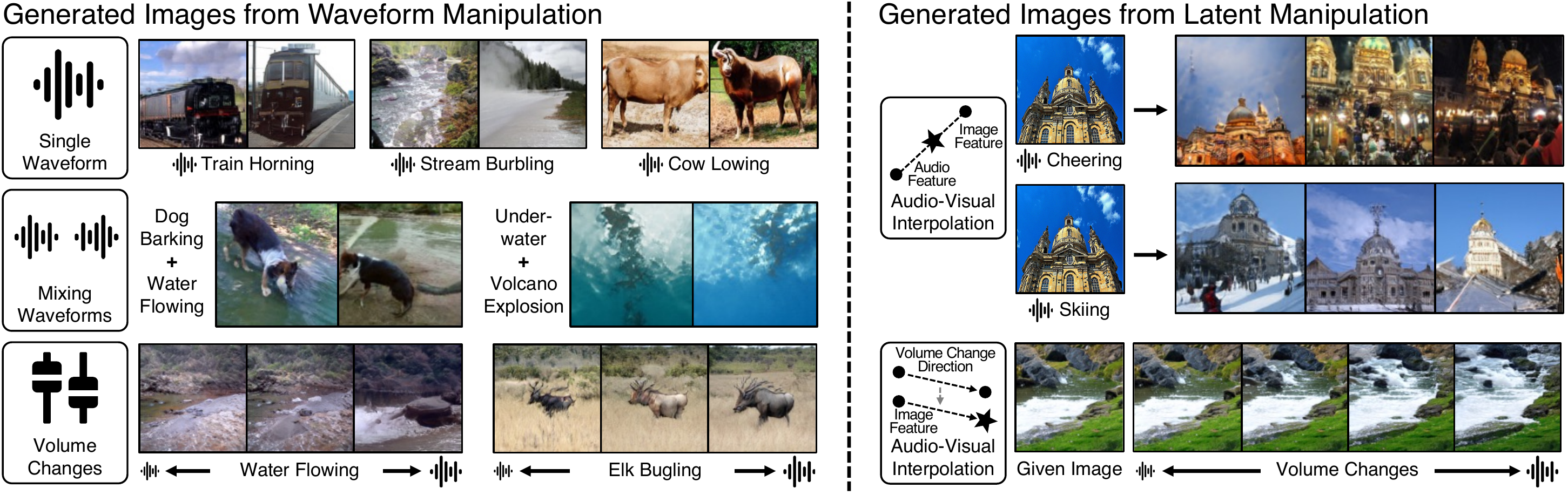}
    \vspace{-6mm}
    \captionof{figure}{{\bf Sound-to-image generation.} We propose a model that synthesizes images of natural scenes from the sound. Our model is trained solely from paired audio-visual data, without labels or language supervision. Our model's predictions can be controlled by applying simple manipulations to the input waveforms (left), such as by mixing two sounds together or by adjusting the volume. We can also control our model's outputs in latent space, such as by interpolating in directions specified by sound (right).}
    \vspace{4mm}
    \label{fig:teaser}

}]

\begin{abstract}
\vspace{-3.1mm}
How does audio describe the world around us? 
In this paper, we propose a method for generating an image of a scene from sound. Our method addresses the challenges of dealing with the large gaps that often exist between sight and sound.
We design a model that works by scheduling the learning procedure of each model component to associate 
audio-visual modalities despite their information gaps. The key idea is to enrich the audio features with visual information by learning to align audio to visual latent space.
We translate the input audio to visual features, then use a pre-trained generator to produce an image.
To further improve the quality of our generated images, we use sound source localization to select the audio-visual pairs that have strong cross-modal correlations.
We obtain substantially better results on the VEGAS and VGGSound datasets than prior approaches.
We also show that we can control our model's predictions by applying simple manipulations to the input waveform, or to the latent space.
\end{abstract}
\vspace{-2.3mm}
\section{Introduction}\vspace{-3mm}
\label{sec:intro}

{\let\thefootnote\relax\footnote{\textbf{Acknowledgment.} This work was supported by IITP grant funded by Korea government (MSIT) (No.2021-0-02068, Artificial Intelligence Innovation Hub; No.2022-0-00124, Development of Artificial Intelligence Technology for Self-Improving Competency-Aware Learning Capabilities). The GPU resource was supported by the HPC Support Project, MSIT and NIPA.
}}
\thispagestyle{empty}

Humans have the remarkable ability to associate sounds with visual scenes, such as how chirping birds and rustling branches bring to mind a lush forest, and the flowing water conjures the image of a river.
These cross-modal associations convey important information, such as the distance and size of sound sources, and the presence of out-of-sight objects.

An emerging line of work has sought to create multi-modal learning systems that have these cross-modal prediction capabilities, by synthesizing visual imagery from sound~\cite{soundguide, li2022learning,towards_a2s,s2b,chen2017,cmcgan,s2i}.  
However, these existing methods come with significant limitations, such as being limited to simple datasets in which images and sounds are closely correlated \cite{towards_a2s,s2b}, relying on vision-and-language supervision~\cite{soundguide}, and being capable only of manipulating the style of existing images~\cite{li2022learning} but not synthesis. 

Addressing these limitations requires handling several challenges. First, there is a significant modality gap between sight and sound, as sound often lacks information that is important for image synthesis, \eg, the shape, color, or spatial location of on-screen objects. Second, the correlation between modalities is often incongruent, \eg, highly contingent or off-sync on timing. 
Cows, for example, only rarely moo, so associating images of cows with ``moo'' sounds requires capturing training examples with the rare moments when on-screen cows vocalize.

In this work, we propose Sound2Scene, a sound-to-image generative model and training procedure that addresses these limitations, and which can be trained solely from unlabeled videos. 
First, given an image encoder pre-trained in a self-supervised way, we train a conditional generative adversarial network~\cite{icgan} to generate images from the visual features of the image encoder. 
We then train an audio encoder to translate an input sound to its corresponding visual feature, by aligning the audio to the visual space. Afterwards, we can generate diverse images from sound by translating from audio to visual embeddings and synthesizing an image. 
Since our model must be capable of learning from challenging in-the-wild videos, we use sound source localization to select moments in time that have strong cross-modal associations.

We evaluate our model on VEGAS~\cite{vegas} and VGGSound~\cite{vggsound}, as shown in \Fref{fig:teaser}.
Our model can synthesize a wide variety of different scenes from sound in high quality, outperforming the prior arts.
It also provides an intuitive way to control the image generation process by applying manipulations at both the input and latent space levels, such as by mixing multiple audios together or adjusting the loudness.
Our main contributions are summarized as follows:
\begin{itemize}
    \item Proposing a new sound-to-image generation method that can generate visually rich images from in-the-wild audio in a self-supervised way.
    \item Generating high-quality images from the unrestricted diverse categories of input sounds for the first time.
    \item Demonstrating that the  samples generated by our model can be controlled by intuitive manipulations in the waveform space in addition to latent space. 
    \item  Showing the effectiveness of training sound-to-image generation using highly correlated audio-visual pairs.

\end{itemize}

\section{Related Work}
\label{sec:related}

\paragraph{Audio-visual generation}
The audio-visual cross-modal generation field is explored in two directions: vision-to-sound and sound-to-vision generation. 
The vision-to-sound task has been actively researched in instrument/music~\cite{audeo,chen2017,cmcgan} and open-domain generic audio generation~\cite{vegas,regnet,visualsound,tamingsound} perspectives. 
In the opposite direction, early work on sound-to-image investigated only restricted and specialized audio domains, such as instruments~\cite{chen2017,cmcgan,strummingbeat,sound2sight}, birds~\cite{s2b} or speech~\cite{speech2face}. Later, Wan~\etal\cite{towards_a2s} and Fanzeres~\etal\cite{s2i} attempt to alleviate the restrictions on the data domain and generate images by conditioning on sounds from nine categories of SoundNet~\cite{soundnet} and five categories of VEGAS~\cite{vegas}, respectively. 
Although we have a similar goal of generating images from unrestricted sounds, 
our approach is capable of handling much more diverse audio-visual generation problems. For example, it is capable of generating images from sounds that come from a variety of categories in the VGGSound~\cite{vggsound} and the VEGAS dataset. Unlike the low-quality results of the previous methods, our model generates visually plausible images that are related to the given audio.

\paragraph{Audio-driven image manipulation} Instead of 
directly generating images from audio, a recent line of work has proposed to edit existing images using sound-based input. Lee~\etal~\cite{soundguide} used the text-based image manipulation model~\cite{styleclip} and extend its embedding space to that of audio-visual modality with the text modality.
Similarly, Li~\etal~\cite{li2022learning} used conditional generative adversarial networks (GANs)~\cite{gan} to edit the visual style of an image to match a sound, and showed that the manipulations could be controlled by adjusting a sound's volume or by mixing together multiple sounds. Our work differs from them in two ways. First, our model is capable of \emph{generating} images conditioned on sound, rather than only \emph{editing} them. Second, unlike Lee~\etal, we do not require a text-based visual-language embedding space. Instead, our model is trained entirely on unlabeled audio-visual pairs.

\paragraph{Cross-modal generation}
Learning to translate one modality to another, \ie, cross-modal generation, is an interesting yet open research problem.
Various tasks have been tackled in diverse domains, such as text-to-image/video~\cite{dalle,dalle2,styleclip,cogview, text2video1, text2video2, text2video3}, speech-to-face/gesture~\cite{speech2face,speech2gesture}, scene graph/layout-to-image~\cite{layout2image,scene2image}, image/audio-to-caption~\cite{sungbin, clipcap, flamingo}, \etc
For bridging the heterogeneous modalities in cross-modal generations, several works~\cite{speech2face,dreamfusion} leverage existing pre-trained models or extend pre-trained CLIP~\cite{clip} embedding space anchored with text-visual modality to suit their purpose~\cite{styleclip,soundguide,dalle2, clipactor}. In this trend, we tackle the task of generating images from sound by leveraging freely acquired audio-visual signals from the video.

\paragraph{Audio-visual learning}
The natural co-occurrence of audio-visual cues is often leveraged as a self-supervision signal to learn the associations between two modalities and assist each other for learning better representations. 
The learned representations in such a way are exploited for 
diverse applications including, 
cross-modal retrieval~\cite{objects,owens2018learning}, video recognition~\cite{chen2021distilling,agreement}, and sound source localization~\cite{less,senocak2018learning,localize19,hardway,park2022marginnce}.
A line of work for constructing an audio-visual embedding space is to jointly train two different neural networks for each modality by judging if the frame and audio correspond to each other~\cite{look,owens18}.
Recent works use clustering~\cite{hu2019deep,alwassel2020self} or contrastive learning~\cite{morgado2021robust,agreement,vatt} for better learning of the joint audio-visual embedding space.
While the above-mentioned approaches learn the audio-visual representations jointly from scratch, another line of work creates a joint embedding space by exploiting existing knowledge from the expert models. The knowledge can be transferred from the audio to visual representation~\cite{owens2016ambient}, visual representation to audio~\cite{soundnet,gan2019self}, or distilled from both audio-visual representations to that of video~\cite{chen2021distilling}.
Our work takes the latter line, assuming a visual expert model exists. We use the image feature extractor to distill rich visual information from large-scale internet videos to the audio modality.

\begin{figure*}[tp]
    \centering
    \includegraphics[width=\linewidth]{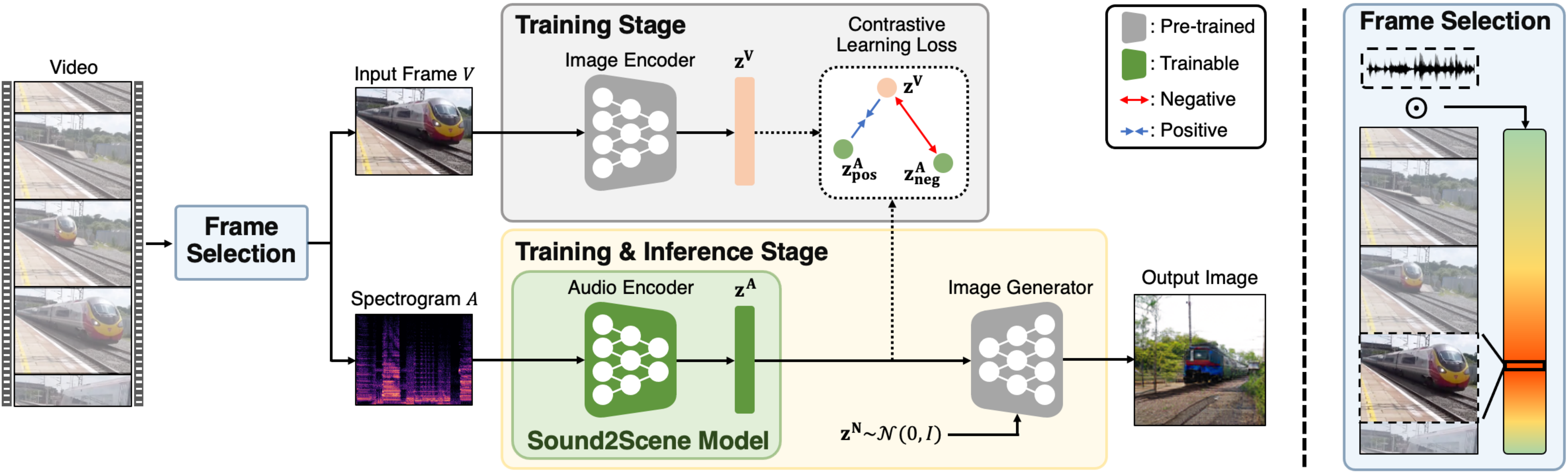}\vspace{-2mm}
    \caption{{\bf Sound2Scene framework.} The frame selection method selects the highly correlated frame-audio segment from a video for training. Then, we train Sound2Scene to produce an audio feature that aligns with the visual feature extracted from the pre-trained image encoder. In the inference stage, the extracted audio feature from input audio is fed to the image generator to produce an image.}
    \vspace{-2mm}
    \label{fig:pipeline}
\end{figure*}

\vspace{-2mm}\section{Method}
\label{sec:method}
The goal of our work is to learn to translate sounds into visual scenes.
Most of the existing methods~\cite{towards_a2s,chen2017,s2i,cmcgan} train GANs to directly generate images from the raw sound or sound features. 
However, the aforementioned challenges and the large variability of visual scenes make the task of directly predicting images from sound challenging.

In contrast to prior approaches, we sidestep these challenges by breaking down the task into sub-problems.
Our proposed Sound2Scene pipeline is illustrated in \Fref{fig:pipeline}. It is composed of three parts: an audio encoder, an image encoder, and an image generator.
First, we pre-train a powerful image encoder and a generator conditioned by the encoder, separately with a large image dataset alone. Since there is a natural correspondence between sound and visual information, we exploit this natural alignment and transfer the discriminative and expressive visual information from the image encoder into audio representation. In this way, we construct a joint audio-visual embedding space that is trained in a self-supervised manner using only in-the-wild videos. Later, the audio representation from this aligned embedding space is fed into the image generator to produce images corresponding to the input sound.

\vspace{-1mm}\subsection{Learning the Sound2Scene Model} \label{ssec:training}
Using the audio-visual data pairs $\mathcal{D}=\{V_i, A_i\}^N_{i=1}$, where $V_i$ is a video frame, and $A_i$ is audio, our objective is to learn the audio encoder to extract informative audio features $\mathbf{z^A}$ that are aligned well with anchored visual features $\mathbf{z^V}$.
Specifically, given the unlabeled data pairs $\mathcal{D}$, the audio encoder $f_{A}(\cdot)$, and the image encoder $f_{V}(\cdot)$, we extract audio features $\mathbf{z^A}{=}f_{A}(A)$ and visual features $\mathbf{z^V}{=}f_{V}(V)$, where $\mathbf{z^V},\mathbf{z^A}\in R^{2048}$. 
Since we exploit the well pre-trained image encoder $f_{V}(\cdot)$, the visual feature $\mathbf{z^V}$ serves as the self-supervision signal for the audio encoder to predict the informative feature $\mathbf{z^A}$ in the way of feature-based knowledge distillation~\cite{hinton, kdsurvey}. 
These aligned features across modalities construct the shared audio-visual embedding space on which the image generator $G(\cdot)$ is separately trained compatibly.

To align the embedding spaces defined by  the heterogeneous modalities, a metric learning approach can be used.
Representations are aligned if they are close to each other under some distance metric. 
A simple approach to align the features of $\mathbf{z^A}$ and $\mathbf{z^V}$ is to minimize the $L_2$ distance, $\norm{\mathbf{z^V}-\mathbf{z^A}}_2$. 
However, we discover that solely using $L_2$ loss can only teach the relationship between two different modalities within the pair without considering the other unpaired samples. 
This results in unstable training and leads to poor image quality.
Therefore, we use InfoNCE~\cite{infonce} as a specific type of contrastive learning, which has been successfully applied to audio-visual representation learnings~\cite{afouras2020AVObjects,hardway,arda2022, chen2021distilling, wav2clip, soundguide}:
\begin{equation}
\texttt{InfoNCE}(\ba_j, \{\bb\}_{k=1}^N) = -\log{\tfrac{\exp(-d(\ba_j,\bb_j))}{\sum^N_{k=1}\exp(-d(\ba_j,\bb_k))}},
\end{equation}
where $\ba$ and $\bb$ denotes arbitrary vectors with the same dimension, and $d(\ba,\bb)=\norm{\ba-\bb}_2$. With this loss, we maximize the feature similarity between an image and its true audio segment (positive) while minimizing the similarity with the randomly selected unrelated audios (negatives).
Given the $j$-th visual and audio feature pair, we first define our audio feature-centric loss
as $L_{j}^{A} = \texttt{InfoNCE}(\mathbf{\hat{z}^A_j}, \{\mathbf{\hat{z}^V}\})$,
where $\mathbf{\hat{z}^A}$ and $\mathbf{\hat{z}^V}$ are representations with unit-norm.
To make our objective symmetric, we compute the visual feature-centric loss term as
$L_{j}^{V} = \texttt{InfoNCE}(\mathbf{\hat{z}^V_j}, \{\mathbf{\hat{z}^A}\})$.
Then, our final learning objective is to minimize the sum of each loss term for all the audio and visual pairs in the mini-batch $B$:
\begin{equation}\label{loss1}
    L_{total} = \textstyle\frac{1}{2B}\sum\nolimits_{j=1}^B\left(L_j^A+L_j^V\right).
\end{equation}

After training the audio encoder with \Eref{loss1}, our model learns visually enriched audio features that are aligned with the visual features. 
Thus, we can directly feed the learned audio feature $\mathbf{z^A}$ with noise vector $\mathbf{z^N}\sim \mathcal{N}(0,I)$ to the frozen image generator as $G(\mathbf{z^N}, \mathbf{z^A})$ to generate a visual scene at the inference stage.

\subsection{Architecture}
All the following modules are separately trained according to the proposed steps.

\paragraph{Image encoder $f_V(\cdot)$}
We use ResNet-50~\cite{he2016deep}. To cope with general visual contents, we train the image encoder in a self-supervised way~\cite{swav} with ImageNet~\cite{imagenet} without labels.

\paragraph{Image generator $G(\cdot)$}
We use the BigGAN~\cite{biggan} architecture to deal with high-quality generation and a large variability of scene contents. 
To make the BigGAN a conditional generator, we follow the modification of the input condition structure of ICGAN~\cite{icgan}.
We train the generator to generate photo-realistic $128\times128$ resolution images from the conditional visual embeddings $\mathbf{z^V}$ obtained from the image encoder.
To train the generator, we use ImageNet without labels in a self-supervised way. While training the image generator, the image encoder is pre-trained and fixed.

\paragraph{Audio encoder $f_A(\cdot)$}
We use ResNet-18, which takes the audio spectrogram as input.
After the last convolutional layer, adaptive average pooling aggregates temporal-frequency information into a single vector.
The pooled feature is fed into a single linear layer to obtain an audio embedding $\mathbf{z^A}$.
The audio network is trained on either VGGSound or VEGAS with the loss in \Eref{loss1} according to target benchmarks.
\begin{figure}[tp]
    \centering
    \includegraphics[width=\linewidth]{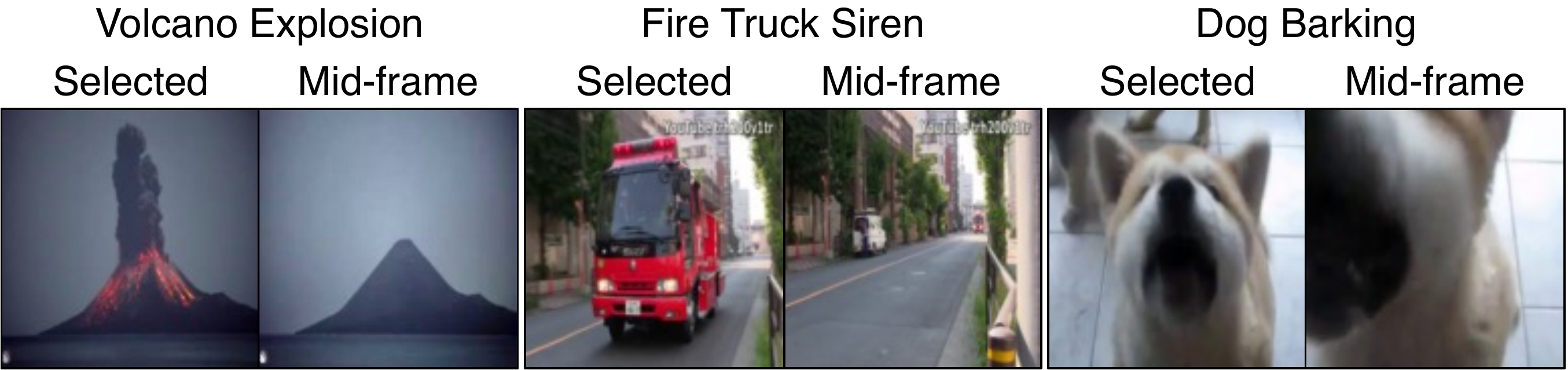}
    \caption{\textbf{Examples of selected \texttt{top-1} frame vs. mid-frame.}}
    \vspace{-2mm}
    \label{fig:pair}
\end{figure}

\subsection{Audio-Visual Pair Selection Module} \label{sssec:data}
Learning the relationship between the images and sounds accurately requires highly correlated data pairs of two modalities. 
Knowing which frame/segment in the video is informative for audio-visual correspondence is not an easy task. 
One straightforward way to collect data pairs for training, $\mathcal{D}$, is to extract a mid-frame of the video with the corresponding audio segment~\cite{soundguide,hardway}. 
However, the 
mid-frame cannot guarantee to contain informative corresponding audio-visual signals~\cite{less}. 
To this end, we leverage a pre-trained sound source localization model~\cite{less} and extract highly correlated audio and visual pairs. 
The backbone networks of~\cite{less} enable us to have fine-grained temporal time steps of audio-visual features, $\mathbf{q^A}$ and $\mathbf{q^V}$, respectively. Correlation scores are computed by $\mathbb{C}_{av}[t]=\mathbf{q^V_t}\boldsymbol{\cdot} \mathbf{q^A_t}$ at each time step.
After computing the correlation scores, $\mathbb{C}_{av}$ are sorted by \texttt{top-k}($\mathbb{C}_{av}[t]$). 
With this correlated pair selection method, we annotate \texttt{top-1} moment frames for each video in the training splits and use them for training.~\Fref{fig:pair} shows the comparison between selected frames and mid-frames. Even though frames are selected automatically, they contain the corresponding distinctive objects to the audio accurately.

\vspace{-2mm}\section{Experiments} \label{sec:exp}
We validate our proposed sound-to-image generation method with experiments on VGGSound~\cite{vggsound} and VEGAS~\cite{vegas}. 
First, we visualize samples of generated images on diverse categories of sounds.
Then, we quantitatively examine the generation quality, diversity, and correspondence between the audio and generated images.
Note that we do not use any class information during training and inference. 
\subsection{Experiment Setup}
\paragraph{Datasets}
We train and test our method on VGGSound~\cite{vggsound} and VEGAS~\cite{vegas}. 
VGGSound is an audio-visual dataset containing around 200K videos. 
We select 50 classes among this dataset and follow the train and test splits provided. VEGAS contains about 2.8K videos with 10 classes. 
For the data statistic balance, we select 800 videos for training and 50 videos for testing per class. 
Test splits in both datasets are used for the following qualitative and quantitative analysis.

\paragraph{Evaluation metrics}
We demonstrate the objective and subjective metrics to evaluate our method quantitatively.
\begin{itemize}
    \item \textbf{CLIP}~\cite{clip} \textbf{retrieval} : Inspired by the CLIP R-Precision metric~\cite{clipr}, we quantify the generated images by measuring image-to-text retrieval performance with recall at $K$ ($R@K$). We feed the generated images and the texts from the name of the audio category to CLIP. Then, we measure the similarities between the image and text features and rank the candidate text descriptions for the query image.
    \item \textbf{Fr\'{e}chet Inception Distance (FID)}~\cite{fid} \textbf{and Inception Score (IS)}~\cite{is} : FID measures the Fr\'{e}chet distance between the features obtained from real and synthesized images using a pre-trained Inception-V3~\cite{inceptionv3}. This same model can also be used for measuring inception score (IS), which computes the KL-divergence between the conditional class distribution and the marginal class distribution. 
    \item \textbf{Human evaluations} : We recruit 70 participants to analyze the performance of our method from a human perception perspective. We first compare our method with the image-only model~\cite{icgan} and then evaluate whether our model generates proper images corresponding to input sound. More details can be found in \Sref{ssec:quan}.
\end{itemize}

\paragraph{Implementation details}
The input of the audio encoder is $1004{\times}257$-dimensional log-spectrogram converted from 10 seconds of audio. 
The extracted frame from the video is resized to $224{\times}224$ and fed as an input of the image encoder. 
We train our model on a single GeForce RTX 3090 for 50 epochs with early stopping. 
We use the Adam optimizer and set the batch size to 64, the learning rate to $10^{-3}$, and the weight decay to $10^{-5}$.

\begin{figure}[tp]
    \footnotesize
    \centering
    \includegraphics[width=\linewidth]{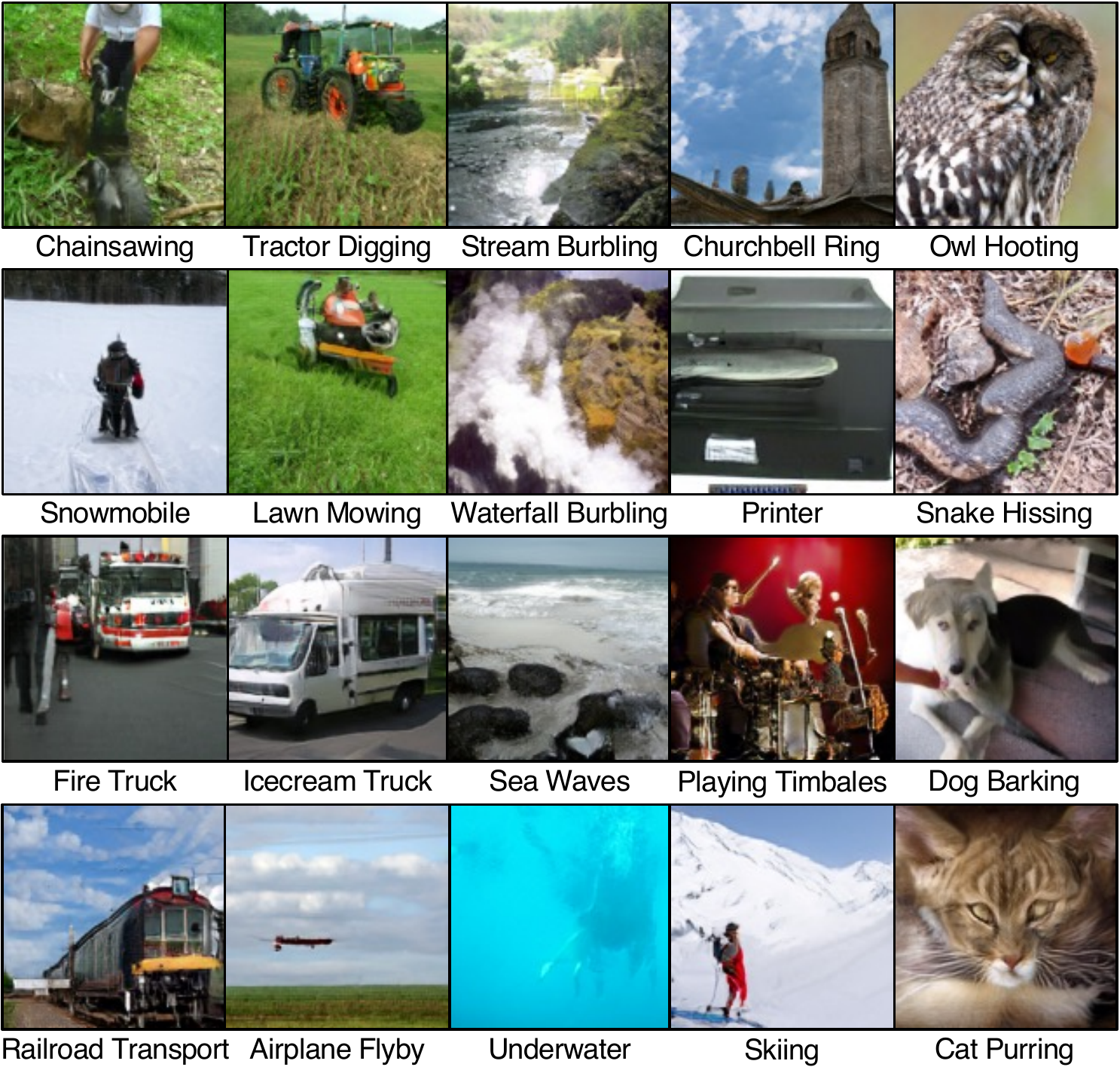}
    \vspace{-6mm}
    \caption{\textbf{Qualitative results by feeding single waveform from VGGSound test set.} Sound2Scene generates diverse images in a wide variety of categories from generic sounds as input.}
    \vspace{-4mm}
    \label{fig:single}
\end{figure}

\subsection{Qualitative Results}
Sound2Scene generates visually plausible images compatible with a single input waveform, as shown in Fig.\ref{fig:teaser} and \ref{fig:single}.
It is not limited to a handful number of categories but handles diverse categories, from animals, and vehicles, to sceneries, \etc  
We highlight that the proposed model can even distinguish subtle differences in similar sound categories, such as ``engine'' (\Fref{fig:single} col. 1-2) and ``water'' (\Fref{fig:single} col. 3) related sounds, and produces accurate and distinct images. See the supplement for more results.

\begin{table*}[t]
\footnotesize
\centering
    \resizebox{0.67\linewidth}{!}{
    \begin{tabular}{ll @{\quad}cc@{\quad}cccc@{\quad}cc}
    \toprule
    &\multirow{2}{*}{Method}& \multirow{2}{*}{\begin{tabular}[c]{@{}c@{}}Encoder\\ ($V$/$A$)\end{tabular}}&  \multirow{2}{*}{\begin{tabular}[c]{@{}c@{}}Generator\\ ($G$/$R$)\end{tabular}}&\multicolumn{4}{c}{VGGSound (50 classes)}&\multicolumn{2}{c}{VEGAS}\\
    \cmidrule(r{4mm}){5-8} \cmidrule{9-10}
     &&&& R@1 & R@5 & FID ($\downarrow$) & IS ($\uparrow$) & R@1& R@5 \\
    \cmidrule{1-10}
    (A)&ICGAN~\cite{icgan} & $V$ & $G$ & 30.06 & 62.59 & \textbf{16.11} & 12.61 &46.60 & 82.48 \\
    (B)&Ours & $A$ & $G$ & \textbf{40.71}  & \textbf{77.36} & 17.97 & \textbf{19.46} & \textbf{57.44} & \textbf{84.08} \\
    \cmidrule{1-10}
    (C)&Retrieval & $A$ & $R$ & 51.28 & 80.37 & - & - & 67.20 & 85.00 \\
    (D)&Upper bound &-&-& 57.82 & 85.79 & - & -&73.60 & 88.2\\
    \bottomrule
    \end{tabular}
    }
    \blank{0.35cm}
    \resizebox{0.30\linewidth}{!}{
    \begin{tabular}{c}
    \includegraphics[width=1\linewidth]{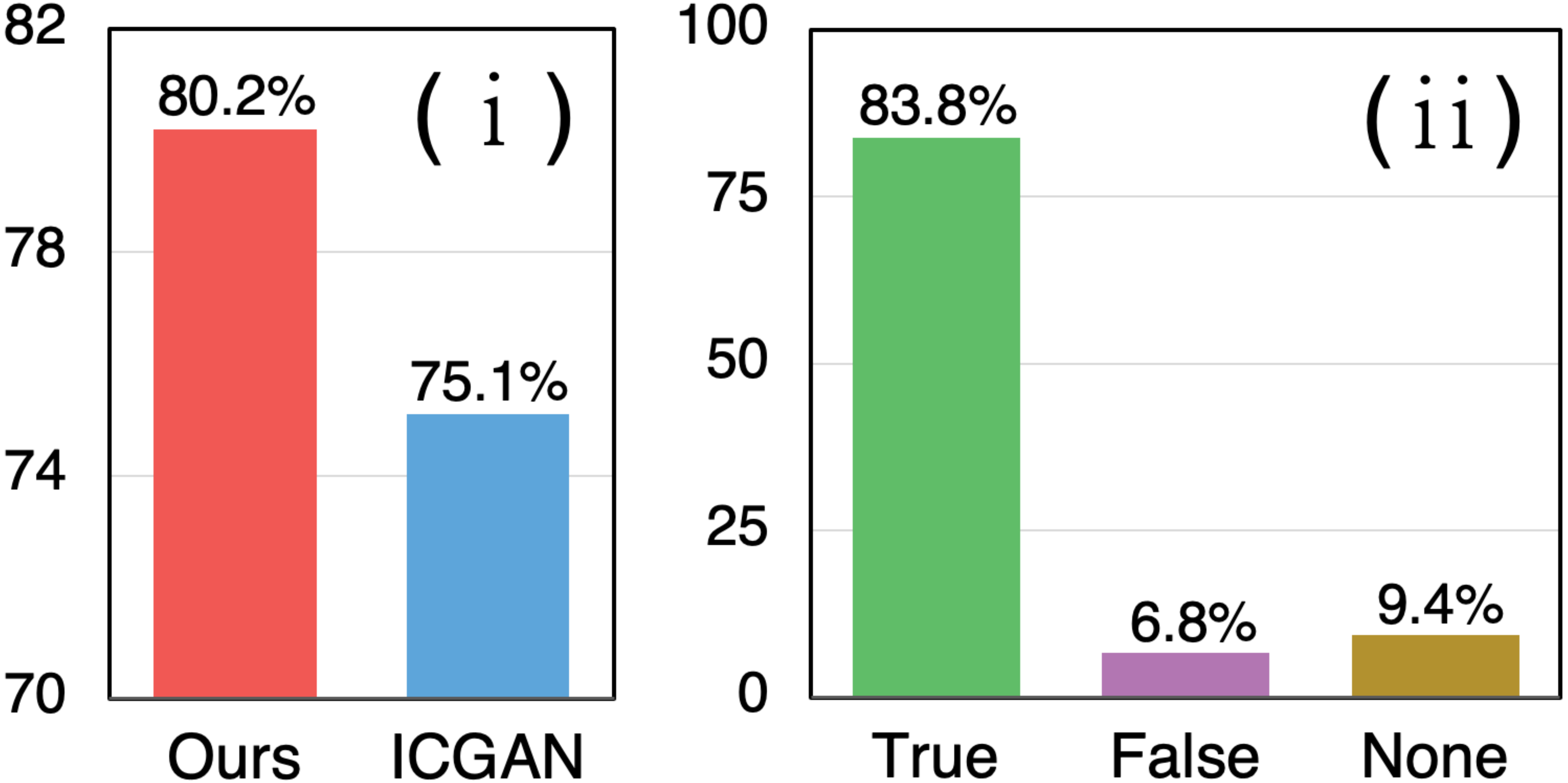}
    \end{tabular}}
    (a) Comparison to baselines \blank{6.3cm} (b) User study\blank{-2.9cm}
    \vspace{-2mm}
   
        \caption{\textbf{Quantitative evaluations.} We compare our method with different baselines (different settings for the encoder and the generator) on CLIP retrieval~(R@k), FID, and IS in (a). 
        For user study, we first compare our method with ICGAN by measuring recall probability between generated images of ICGAN and our method from the same audio-visual pair. Second, we validate our method’s output for the given audio. Results are in (b) respectively.
        $\emph{Abbr.}$ $V$: image encoder, $A$: audio encoder, $G$: image generator, $R$: retrieval system.}
     \vspace{-3mm}
\label{tab:baseline}
\end{table*}

\begin{figure}
    \centering
    \includegraphics[width=\linewidth]{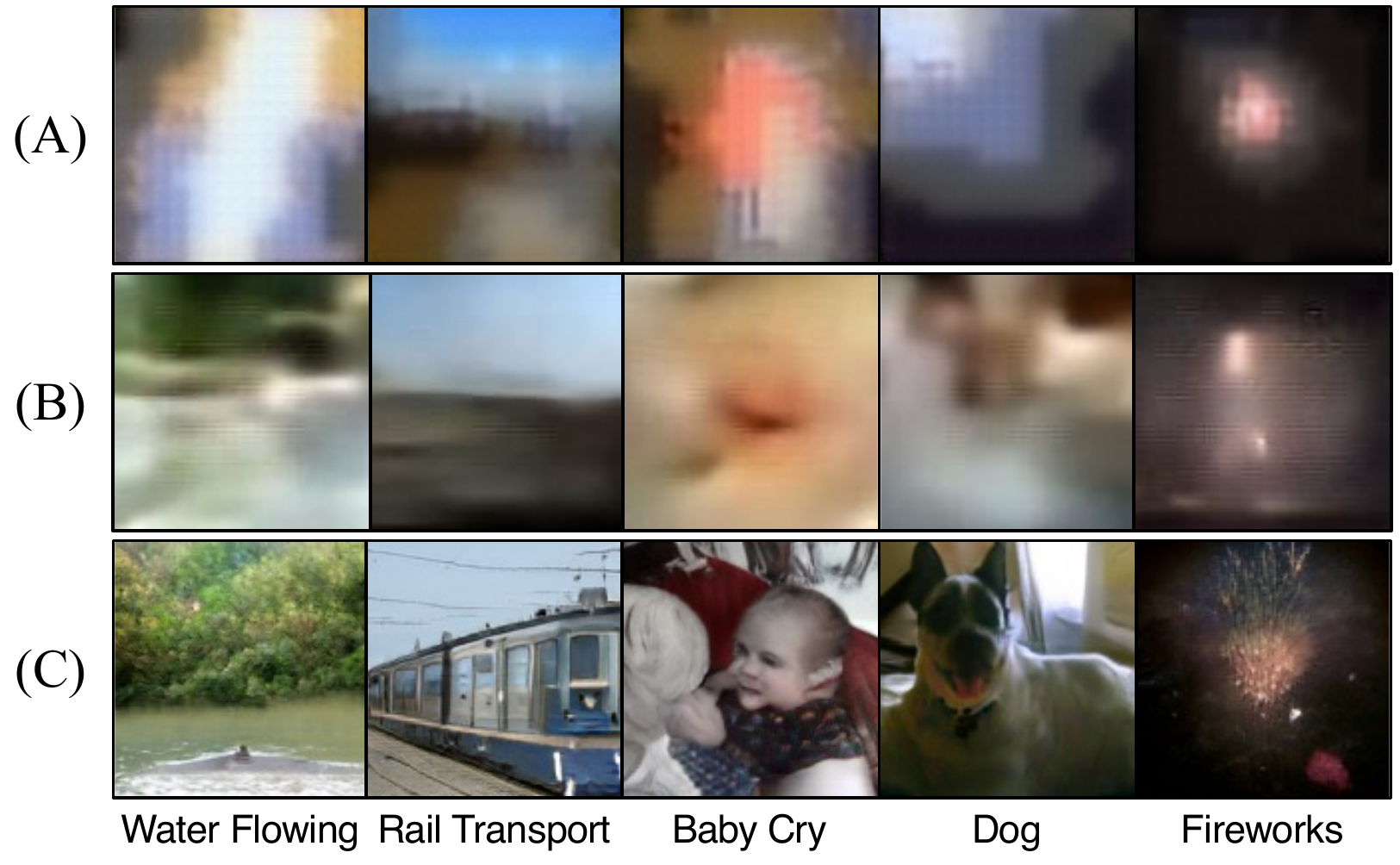}\\
    \vspace{2.5mm}
    \resizebox{0.78  \linewidth}{!}{
    \begin{tabular}{llccc}
    \toprule
    &\multirow{2}{*}{Method}&\multicolumn{3}{c}{VEGAS (5 classes)}\\
    \cmidrule{3-5}
    & & R@1 & FID ($\downarrow$) & IS ($\uparrow$)\\
    \cmidrule{1-5}
    (A)& Pedersoli~\etal~\cite{pedersoli2022estimating} & 23.10 & 118.68 & 1.19\\
    (B)& S2I~\cite{s2i} & 39.19 & 114.84 & 1.45\\
    (C)& Ours & \textbf{77.58} & \textbf{34.68} & \textbf{4.01}\\
    \bottomrule
    \end{tabular}
    }
        \caption{\textbf{Comparison to the baseline}~\cite{pedersoli2022estimating} \textbf{and existing sound-to-image method}~\cite{s2i}. Our method outperforms the others both qualitatively and quantitatively in the VEGAS dataset.}
    \vspace{-4mm}
    \label{fig:competitive}
\end{figure}

\subsection{Quantitative Analysis} \label{ssec:quan}
\paragraph{Comparison with other methods}
We compare our model with the prior arts of which codes are publicly available, S2I\footnote{\url{https://github.com/leofanzeres/s2i}}~\cite{s2i} and Pedersoli\footnote{\url{https://github.com/ubc-vision/audio_manifold}}~\etal~\cite{pedersoli2022estimating}.
Note that Pedersoli~\etal is not targeted for sound-to-image but uses VQVAE-based model~\cite{vqvae} for sound-to-depth or segmentation generations.
Though our model can handle more diverse categories of in-the-wild audio, we follow the training setup as in S2I by training our model and Pedersoli~\etal with five categories in VEGAS for a fair comparison.
As shown in \Tref{fig:competitive}, our model outperforms all the other methods. Additionally, it generates visually plausible images, while previous methods fail to generate recognizable images. 
We postulate that learning visually enriched audio embeddings combined with a powerful image generator leads to superior results.\footnote{More qualitative comparisons can be found in the supp. material.}

\paragraph{Comparison with strong baselines}
We further compare our proposed method with closely-related baselines in~\Tref{tab:baseline}~(a). 
First, we compare with an image-to-image generation model identical to ICGAN~\cite{icgan} (A). Our model (B) shares the same image generator with (A), but differs in the encoder type and the input modality. As shown, (B) outperforms or gives comparable results to (A) in all metrics. We presume that the noisy characteristic of the video datasets causes (A) to fail to extract a good visual feature for image generation, while audio is relatively robust and less sensitive to those limitations, resulting in generating plausible images. We also compare our model (B) with a retrieval system (C) that can be regarded as a strong baseline. Given an input audio embedding, the retrieval system finds the closest image from the database. (C) shares the same audio encoder with (B), but the image generator $G$ is replaced with the same memory-sized database of the images from the training data. (D) is an upper bound in which the extracted video frames are directly used for the evaluations.
The performance gap between (C) and (D) is dramatically lower than that of (B) and (D), which justifies that our audio encoder properly maps the input audio to the joint embedding space.
(C) outperforms (B) on $R@1$ for both datasets, while (B) performs comparably to (C) in $R@5$.
Though the image generators have room to improve, these results show that our method can reach to the proximity of the strong baseline.

\begin{table}[t]
\vspace{1.5mm}
\footnotesize
\centering
    \resizebox{0.95\linewidth}{!}{
    \begin{tabular}{l@{\quad}l@{\quad}c@{\quad}c@{\quad}c@{\quad}c@{\quad}c@{\quad}c}
    \toprule
    \multirow{2}{*}{}&\multirow{2}{*}{Loss}&\multirow{2}{*}{$F$}
    &\multirow{2}{*}{Duration}
    &\multicolumn{4}{c}{VGGSound (50 classes)}\\
   
    \cmidrule{5-8}
     & &&& R@1 & R@5 & FID ($\downarrow$) & IS ($\uparrow$)\\
    \cmidrule{1-8}
    (A) &$L_2$ & \checkmark&10 sec.& 18.21 & 46.69 & 24.05 & 9.97\\
    (B) & $L_{nce}$& \checkmark&10 sec.& 31.63 & 66.04 & 27.05 & 12.92\\
    (C) & $L_{total}$& & 10 sec.&37.20 & 73.13 & 21.20 & 17.51\\
    \cmidrule{1-8}
    (D) &\multirow{2}{*}{$L_{total}$}&\multirow{2}{*}{\checkmark}&1 sec. &35.85& 72.02&19.05&17.87\\
    (E) &&&5 sec. &38.24&75.76&20.43&18.81\\
    \cmidrule{1-8}
    (F) & $L_{total}$& \checkmark& 10 sec.& \textbf{40.71} & \textbf{77.36} & \textbf{17.97} & \textbf{19.46}\\
    \bottomrule
    \end{tabular}
    }
    \caption{\textbf{Ablation studies of our proposed method.} We compare the different configurations of our method by changing the loss functions, frame selection method, and duration of the audio. $F$ denotes the frame selection method. 
     }
     \vspace{-2mm}
\label{tab:ablation}
\end{table}

\paragraph{User study}
We summarize the user study in \Tref{tab:baseline}~(b) with two experiments: (\romannumeral1) comparison to ICGAN and (\romannumeral2) validation of the proper image generation for given audio.
Each experiment has 20 questions.
In (\romannumeral1), audio and five images are given to the participants.
Among the five, two are generated by ours and ICGAN, respectively, and the rest are randomly generated from either method.
Participants choose all the images that illustrate the given sound, and we check the preference by comparing the recall probability of ICGAN and ours.
In (\romannumeral2), audio and four images are provided to the participants. 
All four images are generated by ours, but only one is from the given sound.
Participants choose only one image that best illustrates the given sound. As in (\romannumeral1), our model is more preferred. 
Moreover, (\romannumeral2) shows that the precision of our method is 83.8\%, which supports our model generating highly-correlated images to the given sounds.

\paragraph{Ablation studies}
We conduct a series of experiments in order to verify our design choices in \Tref{tab:ablation}.
We compare the performance of applying different distillation losses: a simple $L_2$ loss between the image and audio feature, and InfoNCE loss~\cite{infonce} with a cosine similarity measurement, $L_{nce}$, rather than using $L_2$ distance as in ~\Eref{loss1}. 
As the results of (A), (B), and (F) reveal, our loss choice (F) leads to producing more diverse and higher quality results.
We also observe that the frame selection method, discussed in~\Sref{sssec:data}, brings extra performance improvement; see the performance difference between (C) and (F).
Finally, we test the effect of audio duration. 
We train models with 1, 5, and 10 seconds of sounds with other experimental settings being fixed. 
By comparing (D), (E), and (F), we observe that feeding longer sounds consistently improves performance.
We presume that the longer audios capture more descriptive audio semantics, while shorter ones are vulnerable to missing them.

\section{Controllability of Sound2Scene}
Our model learns the natural correspondence between audio and visual signals via aligned audio-visual embedding space. Thus, intuitively, we ask if manipulations on input can result in corresponding changes in the generated images. We observe that even without an explicit objective, our model allows controllable outputs by applying simple manipulations on inputs in the \emph{waveform space} or learned \emph{latent space}. 
This opens up interesting experiments that we explore below.

\subsection{Waveform Manipulation for Image Generation}
\begin{figure}[tp]
    \centering
    \includegraphics[width=\linewidth]{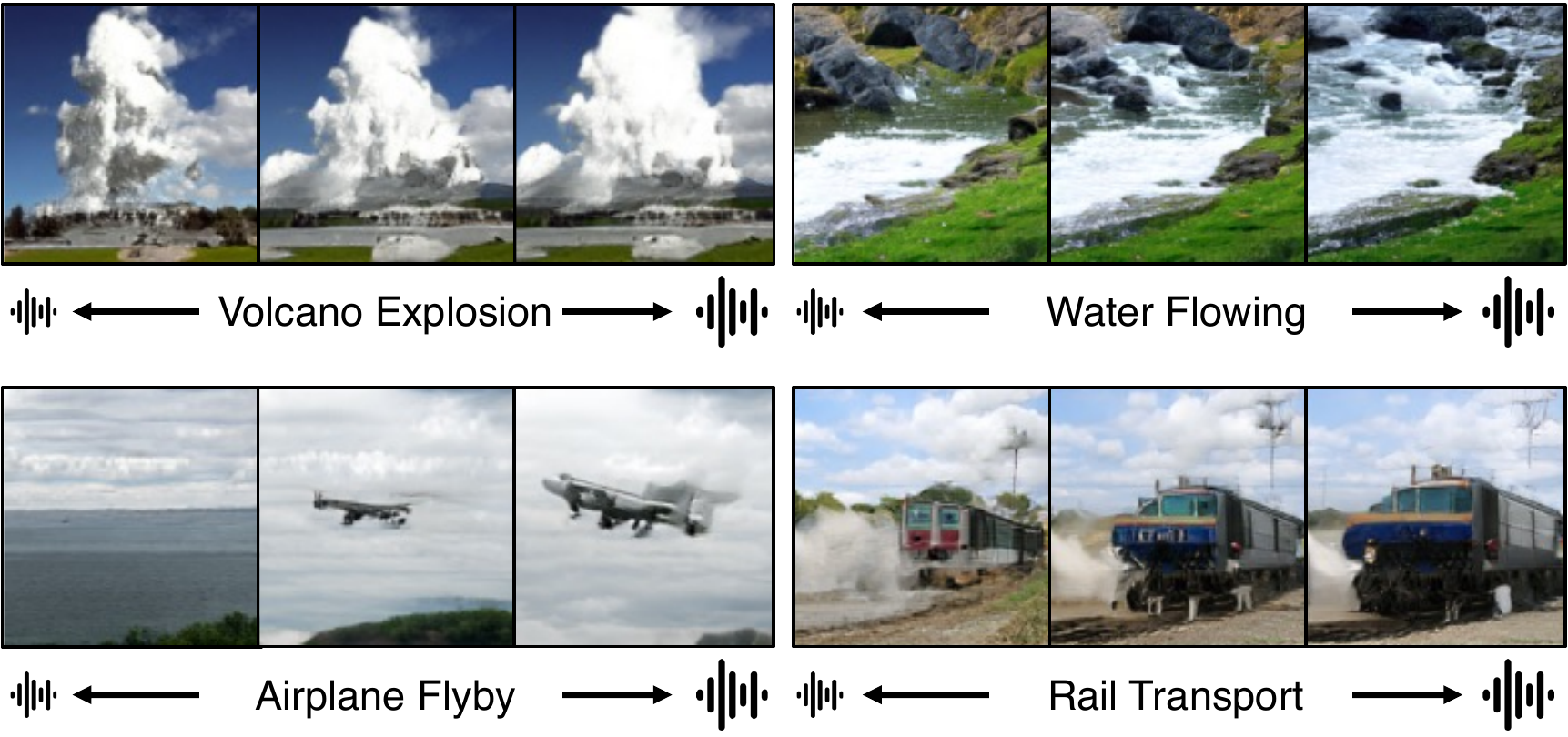}
    \caption{\textbf{Generated images by changing the volumes of the input audio in the \emph{waveform space}.} As the volume increases, the objects of the sound source become larger or more dynamic.}
    \vspace{1.5mm}
    \label{fig:volume}
\end{figure}

\paragraph{Changing the volume}
Humans can roughly predict the distance or the size of an instance by the volume of the sound. 
To check if our model can also understand the volume differences, we reduce and increase the volume of the reference audio. 
Each audio with a different volume is fed into our model with the same noise vector.
As shown in \Fref{fig:volume}, the instances in the synthesized images get larger as the volumes increase.
Interestingly, the volume changes in ``Water Flowing'' illustrate the different flows of water, while ``Rail Transport'' shows a train approaching in the scene. 
These results highlight that our model has not only class-specific understanding but also the relation between the volume of the audio and visual changes.
We assume that the supervision from the visual modality enables our model to capture such strong and expressive audio-visual relationships.

\begin{figure}[tp]
    \centering
    \includegraphics[width=\linewidth]{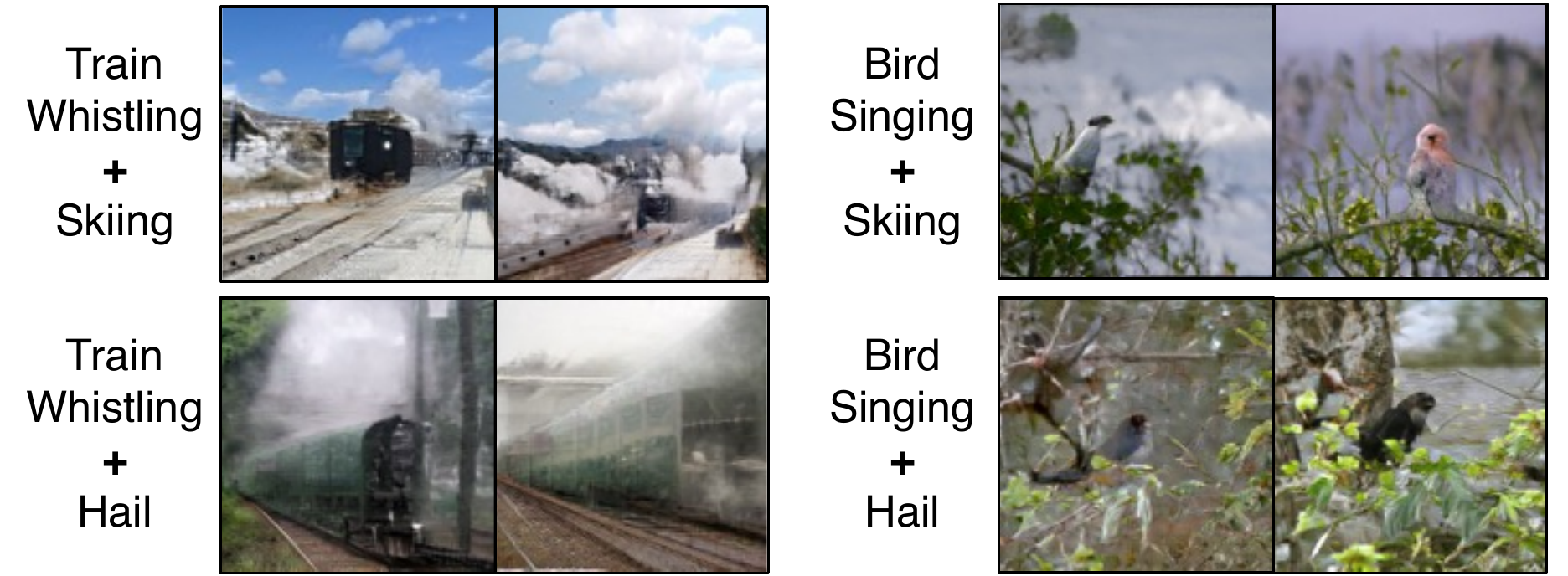}
    \vspace{-6mm}
    \caption{\textbf{Generated images by mixing two different audios in the \emph{waveform space}.}} 
    \vspace{-5mm}
    \label{fig:mix}
\end{figure}

\paragraph{Mixing waveforms}
We investigate if our model can capture the existence of multiple sounds in the generated images.
To this end, we mix two waveforms into a single one and feed it to our model.
As shown in~\Fref{fig:teaser} and~\Fref{fig:mix}, our model synthesizes images by reflecting those multiple audio semantics. 
For example, the railroad or a bird pops up across the snowy scene when mixing with the ``Skiing'' sound, and the train and bird appear in the misty scene when mixing with the ``Hail'' sound. Also, as in~\Fref{fig:teaser}, mixing the ``Dog Barking'' and ``Water Flowing'' sounds outputs a scene with a dog playing in the water.
Sensing multiple separate sounds from a single mixed audio input~\cite{hu2020discriminative}, \ie, audio source separation~\cite{gao2019co}, and generating their visual appearance in the proper context is not trivial.  
However, our results show that the proposed model can handle this to a certain extent.

\begin{figure}[tp]
    \centering
    \footnotesize
    \includegraphics[width=\linewidth]{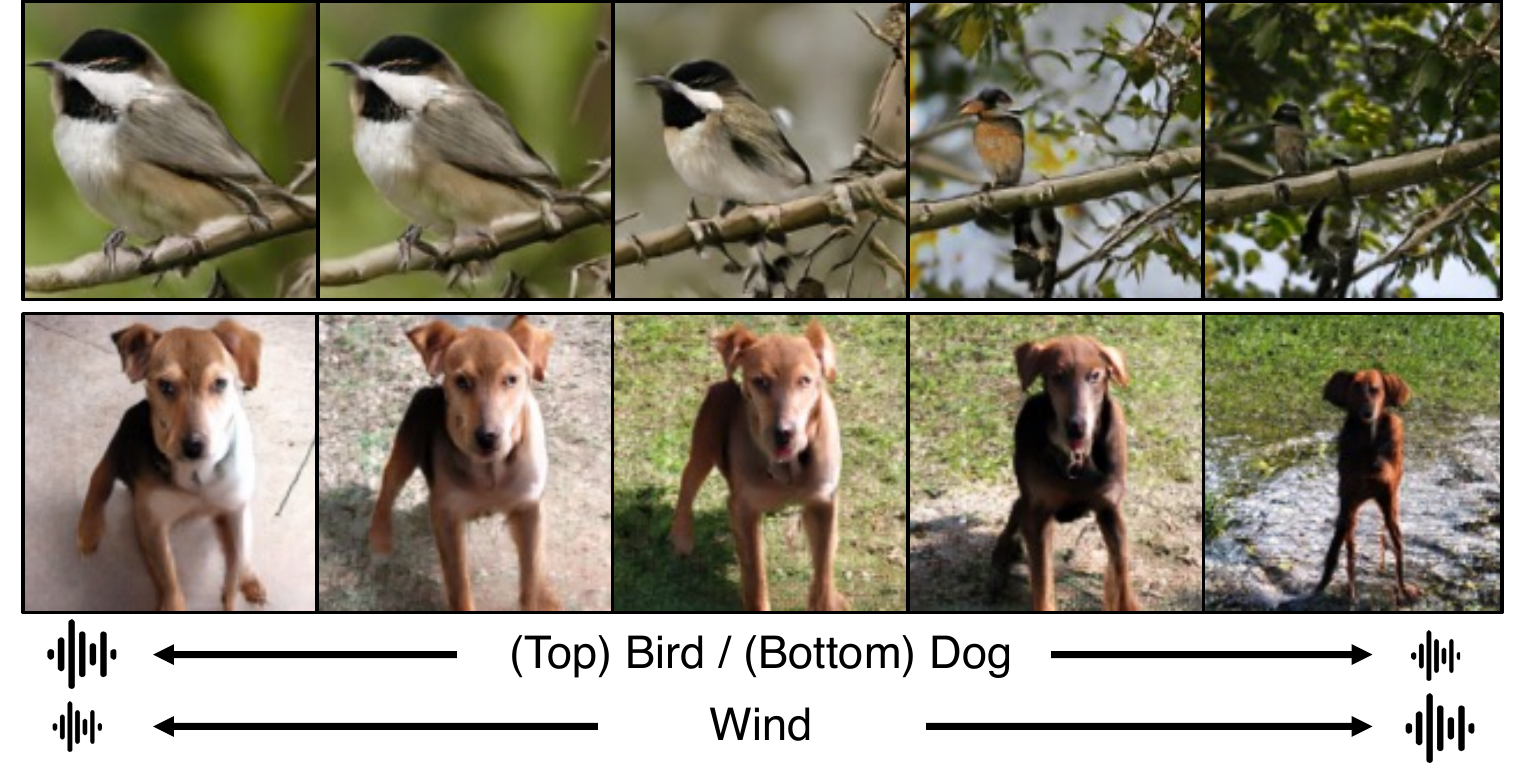}
    \vspace{-6mm}
    \caption{\textbf{Generated images by mixing multiple audios with volume changes in the \emph{waveform space}}. We observe that Sound2Scene mimics the camera movement by placing the object further as the wind sound gets larger.}
    \vspace{-3mm}
    \label{fig:mixvolume}
\end{figure}

\paragraph{Mixing waveforms and changing the volume}
Here, we manipulate the input waveform by combining the multiple waveforms and changing their volumes at the same time.
In \Fref{fig:mixvolume}, we mix the ``Wind'' sound with each of the ``Bird'' and ``Dog'' sound with volume changes.
As the ``Wind'' sound gets larger while the ``Bird'' sound decreases, the bird gets smaller and is finally covered with the bushes.
In the same experiment setting, a close-up shot of the dog indoors starts zooming out and gets a wide shot in the outdoor environment.
These results show that our model can capture subtle changes in the audio and reflect them to generate images.

\subsection{Latent Manipulation for Image Generation}
As we construct an aligned audio-visual embedding space, our model can also take an image and audio together as input and generate images.
We introduce two different approaches (\Fref{fig:teaser}) for audio-visual conditioned image generation where both use the features of the inputs in the latent space.

\begin{figure}[tp]
    \centering
    \small
    \includegraphics[width=\linewidth]{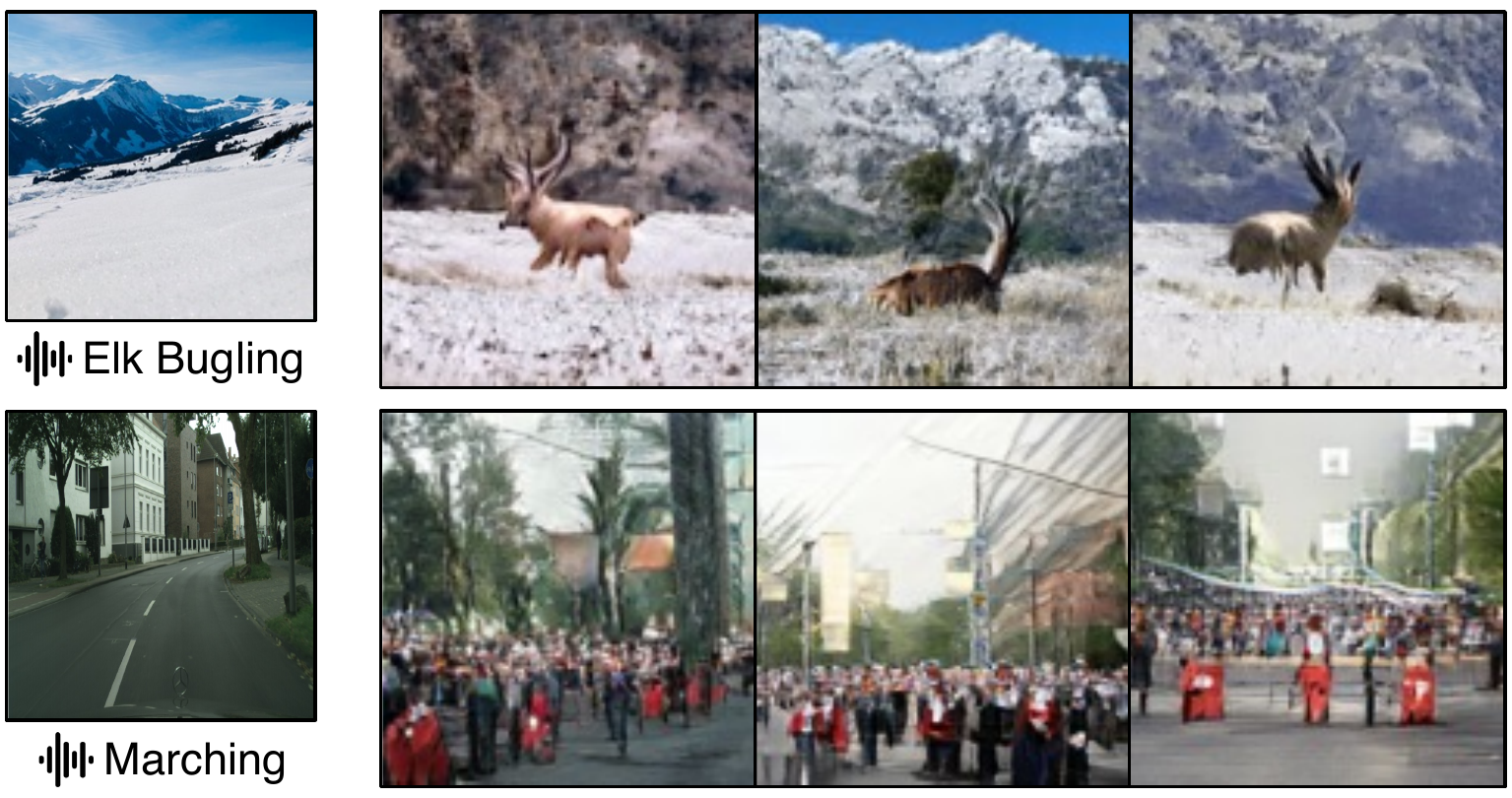}\\
    \blank{-1.4cm}(a) Inputs \blank{2.1cm} (b) Generated images
    \vspace{-2mm}
    \caption{\textbf{Generated images conditioned on image and audio.} We interpolate between a given visual feature and an audio feature in the \emph{latent space}. This interpolated feature is then fed to the image generator to get a novel image.}
    \vspace{-4mm}
    \label{fig:imgaud}
\end{figure}

\begin{figure}[t]
    \centering
    \footnotesize
    \resizebox{0.2\linewidth}{!}{
    \begin{tabular}{c}
    \includegraphics[width=0.4\linewidth]{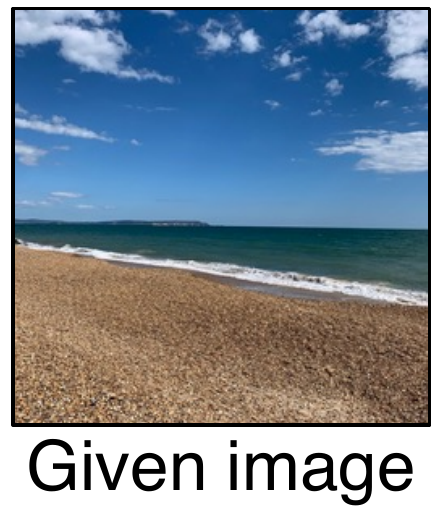}
    \end{tabular}}
    \resizebox{0.7\linewidth}{!}{
    \begin{tabular}{lc}
    \toprule
    Method&Target task\\
    \cmidrule{1-2}
    \cite{soundguide}&Sound-guided image manipulation\\
    Ours&Sound-to-image generation\\
    \bottomrule
    \end{tabular}
    }
    \vspace{1.5mm}
    \\
    \includegraphics[width=\linewidth]{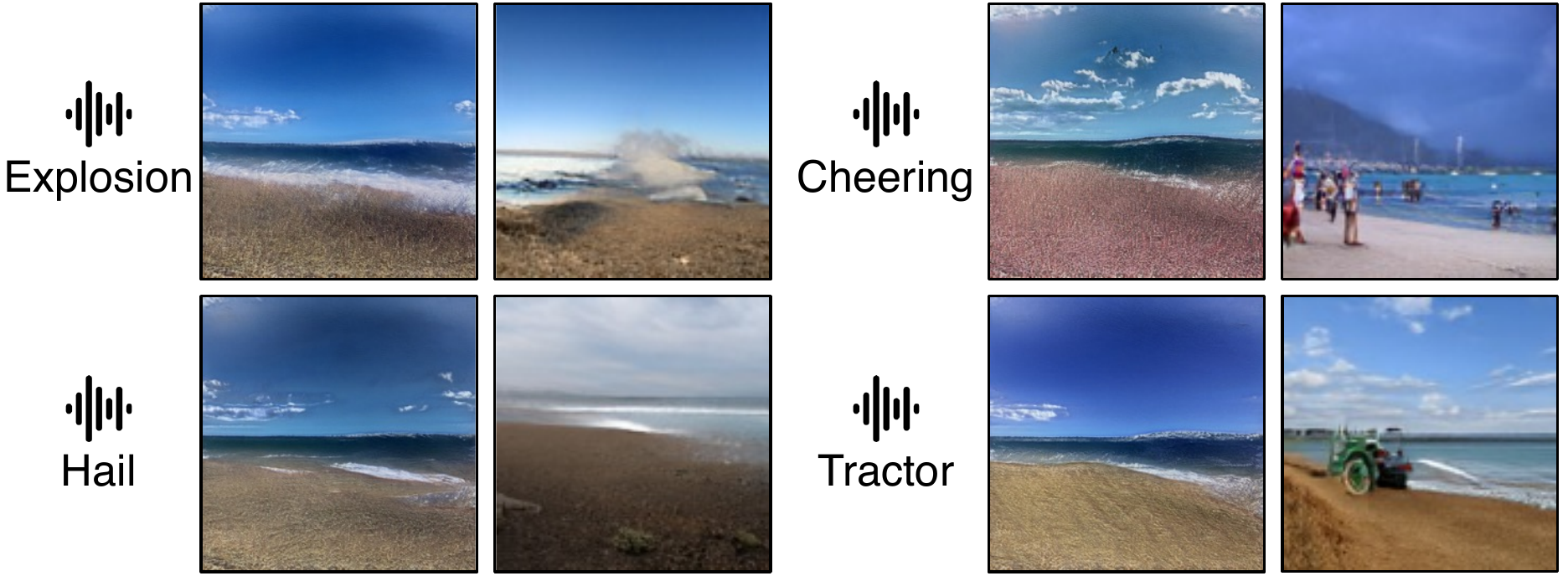}
    \blank{0.56cm} \cite{soundguide} \blank{0.90cm} Ours \blank{2.00cm}\cite{soundguide} \blank{0.90cm} Ours\blank{-0.5cm}
    \vspace{-1mm}
    \caption{\textbf{Qualitative comparison of our method and Lee~\etal}~\cite{soundguide}. Lee~\etal fail to insert an object while maintaining the contents of the given image. Our method, by contrast, successfully inserts objects that sound in the scene by generating a new image. Note that both works target different tasks.}
    \vspace{-4mm}
    \label{fig:korea}
\end{figure}

\paragraph{Image and audio conditioned image generation}
Given an image and audio, we extract a visual feature $\mathbf{z^V}$ and an audio feature $\mathbf{z^A}$.
Then, we interpolate two different features in the latent space and obtain a novel feature: $\mathbf{z^{new}} = \lambda\mathbf{z^V} + (1-\lambda)\mathbf{z^A}$, where $\lambda$ differs throughout the examples.
This feature is fed to the image generator for generating an image.
As shown in~\Fref{fig:imgaud}, this simple approach can produce an image by putting the sound context into the scene, such as a marching sound bringing parade-looking people or a loud elk sound making an elk appears in a snowy scene.

We further use this approach to compare our method with the recent sound-guided image manipulation approach\footnote{\url{https://github.com/kuai-lab/sound-guided-semantic-image-manipulation}}~\cite{soundguide} in~\Fref{fig:korea}. Note that this task is not targeted explicitly by our model but appears as a natural outcome of our design. While Lee~\etal~\cite{soundguide} preserves the overall content of the given image, it fails to insert an object corresponding to the sound. In contrast, our method creates an image (nearly the same as the given one) by conditioning on both modalities, for example, inserting an explosion and tractor in the scene or making the ocean view look cloudy due to the hail.

\begin{figure}[tp]
    \centering
    \includegraphics[width=\linewidth]{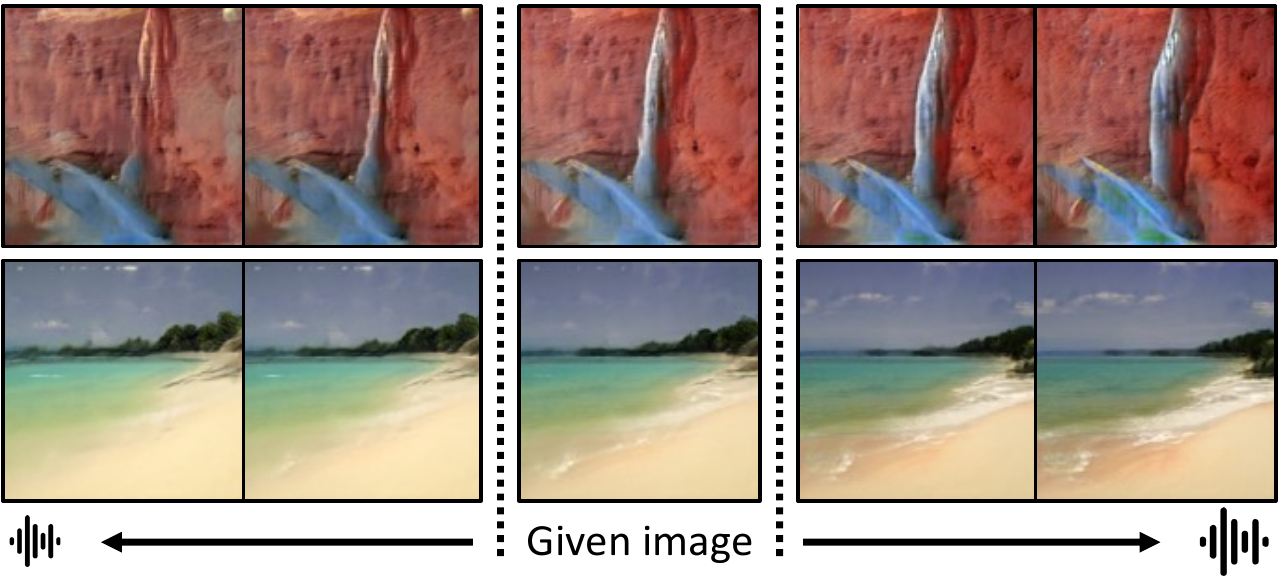}
    \vspace{-7mm}
    \caption{\textbf{Image editing by volume changes in \emph{latent space}.} We extract an image feature and noise vector by GAN inversion, and two audio features with different volumes. Then, we move the image feature in the direction of the audio feature differences.}
    \label{fig:edit}
    
\end{figure}

\paragraph{Image editing with paired sound}
We approach sound-guided image editing from a different perspective by manipulating the inputs in the latent space. By using GAN inversion~\cite{gan_inv1,gan_inv2}, we extract a visual feature $\mathbf{z^V_{inv}}$ and corresponding noise vector $\mathbf{z^N_{inv}}$ for the given image. Additionally, we change the volume of the corresponding audio and extract two different audio features, $\mathbf{z^A_1}$ and $\mathbf{z^A_2}$, respectively. We move the visual feature toward the direction of the difference between the two audio features and obtain a novel feature:
$\mathbf{z^{new}} = \mathbf{z^V_{inv}} + \lambda(\mathbf{z^A_1}-\mathbf{z^A_2}$), \ie, manipulating the visual feature with the audio guidance. By using this new feature with image generator, $G(\mathbf{z^N_{inv}}, \mathbf{z^{new}})$, the original image is edited. As shown in \Fref{fig:edit}, by simply changing the volume -- moving through the latent space -- we can change the flow of the waterfall or make the ocean wave stronger or calmer. 

\section{Discussion}
\label{sec:limitations}

\setlength{\columnsep}{4mm}%
\begin{wrapfigure}{r}{0.21\textwidth}
\centering
\vspace{-4mm}
\includegraphics[width=0.22\textwidth]{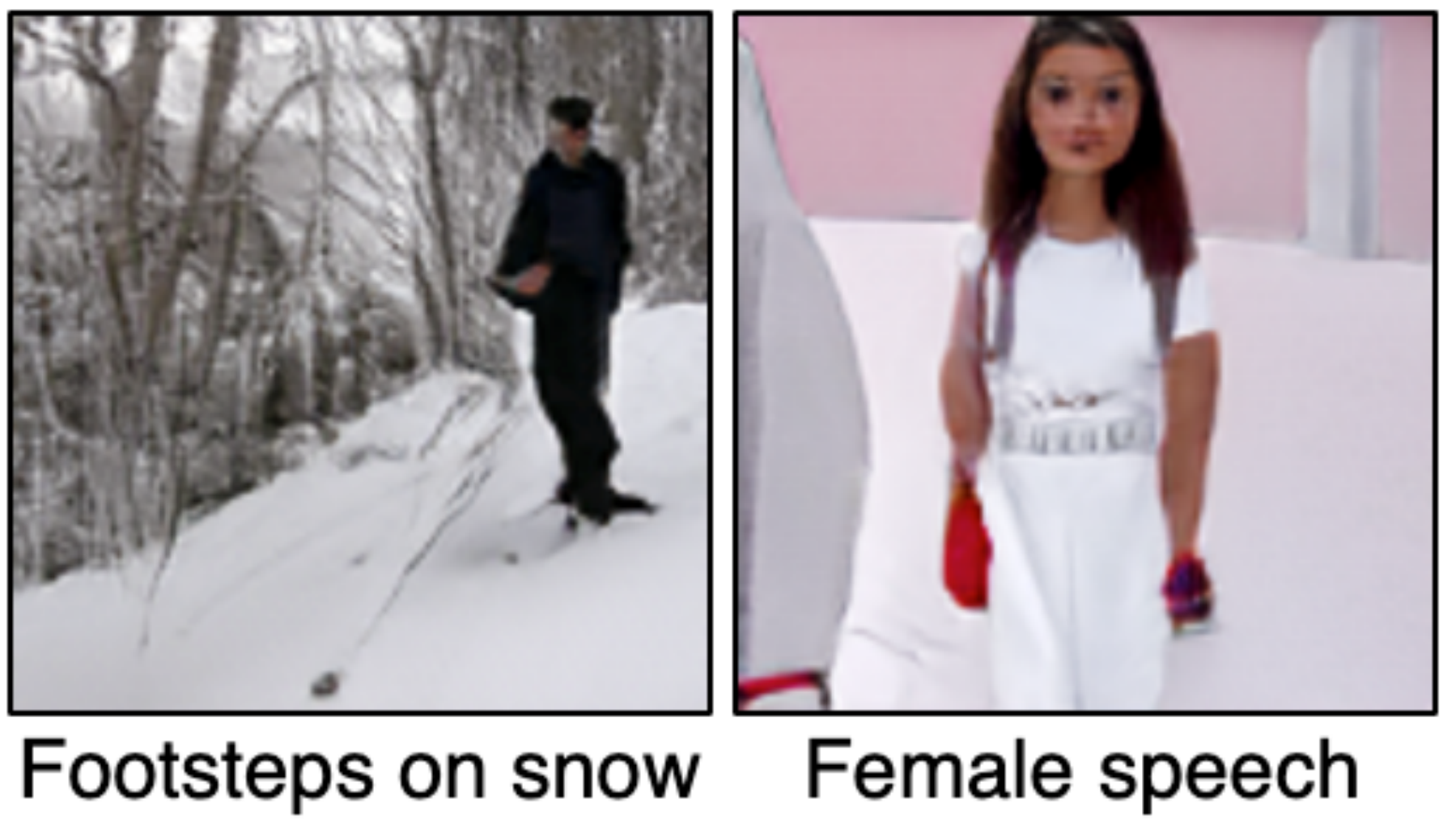}\\
\vspace{-3mm}
\caption{\textbf{Generalization to unseen classes.}}\label{fig:general}
\vspace{-3mm}
\end{wrapfigure}
\paragraph{Generalization} We show generalization of the model    
\noindent to \emph{some extent} in two settings: 1) Generating images from unseen categories that are semantically similar to the training set as sound often carries overlapping information (\Fref{fig:general}). 
2) Compositionality (dog barking+water flowing in~\Fref{fig:teaser}). However, our method may not be generalized for every unseen category as similar limitation is also common in other X-to-Vision tasks.

\begin{figure}[tp]
    \centering
    \footnotesize
    \includegraphics[width=\linewidth]{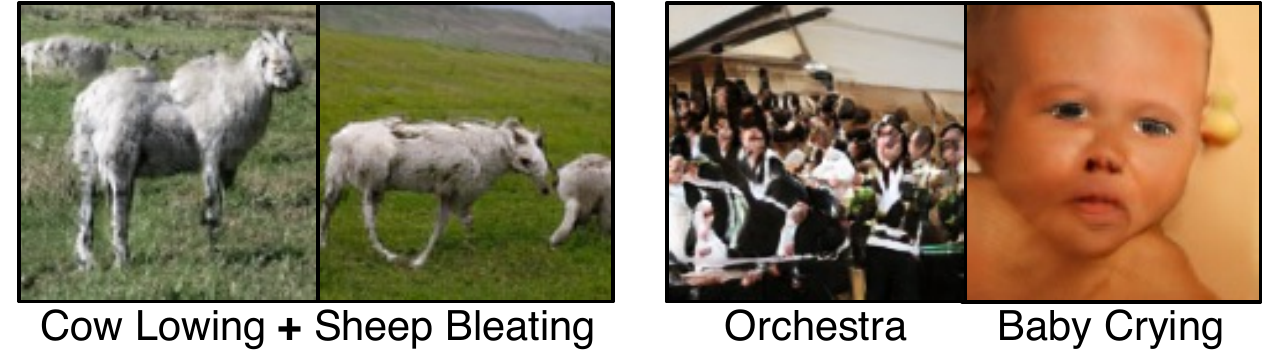}\\
    \blank{0.6cm} (a) Mixing objects \blank{1.4cm} (b) Incomplete human forms
    \vspace{-2mm}
    \caption{\textbf{Examples of failure results.} In cases where both source audio specify certain objects, our model mixes two objects into one (a). Our model also tends to produce incomplete human forms (b).}
     \vspace{-4mm}
    \label{fig:failure}
\end{figure}

\paragraph{Failure cases}
Although our method shows favorable results for given audios, single or mixture, there are some failure cases we have observed.
The first phenomenon is that 
our model often generates images with a single or a blended object
unintentionally when the audios used in the mixture specify two distinct but similar-sounding objects
(see examples in~\Fref{fig:failure} (a)). Another one we have observed is that the quality of the outputs is lower in human-related categories (see examples in~\Fref{fig:failure} (b)). 
This phenomenon is shared across typical GAN-based generative models, \eg, \cite{multigan},  
when training the generator with generic object dataset, \ie, ImageNet.

\paragraph{Conclusion}
In this paper, we propose Sound2Scene, a model for generating images that are relevant to the given audio. This task inherently has challenges: a significant modality gap between audio and visual signals, such that audio lacks visual information, and audio-visual pairs are not always correspondent. Existing approaches have limitations due to these difficulties. We show that our proposed method overcomes these challenges in that it can successfully enrich the audio features with visual knowledge, selects audio-visually correlated pairs for learning, and generates rich images with various characteristics. Furthermore, we demonstrate our model allows controllability in inputs to get more creative results, unlike the prior arts. We would like to note that our proposed learning approach and the audio-visual pair selection method are independent of the specific design choice of the model. We hope that our work encourages further research on multi-modal image generation.

\newpage
{\small
\bibliographystyle{ieee_fullname}
\bibliography{11_references}
}

\ifarxiv \clearpage \renewcommand{\thefigure}{S\arabic{figure}}
\setcounter{figure}{0}

\appendix
\label{sec:appendix}
\noindent {\LARGE \textbf{Appendix}}
\vspace{6mm}

The contents in this appendix are as follows: A. Implementation details (\Sref{sec:a}), B. Suitable categories for sound-to-image (\Sref{sec:b}), C. Comparison with the prior arts (\Sref{sec:d}),
D. Additional qualitative analysis and results (\Sref{sec:e}), and E. Details of the user study (\Sref{sec:f}).
We also recommend watching the supplementary video, containing generated images and corresponding input sounds.

\section{Implementation Details}\label{sec:a}
\paragraph{Audio pre-processing}
The input for the audio encoder is $1004\times257$-dimensional log-spectrogram, converted from 10 seconds of audio. 
We first extract up to 10 seconds of audio from the beginning of each video. 
If the video clip is shorter than 10 seconds, we repeat the audio to have the expected input length.
Then, we resample the audio waveform at 16kHz and convert it into frequency domains by constructing a spectrogram. 
The spectrogram is passed through a logarithm function before using it as input.

\paragraph{Evaluation metric}
In the main text, we use Fr\'{e}chet Inception Distance (FID)~\cite{fid} and Inception Score (IS)~\cite{is} metrics to evaluate the quality and diversity of the generated images. 
To measure both of the metrics, the Inception-V3~\cite{inceptionv3} model is required.
We fine-tune the Inception-V3 model on VGGSound~\cite{vggsound} and compute FID and IS with 30k generated images from the test set.

\section{Suitable Categories for Sound-to-Image}\label{sec:b}
Not every category is suitable to be used to infer visual scenes from sounds. In this section, we analyze which categories are not only audio-visually well-corresponded but also suitable for sound-to-image translation.

As described in~\cite{winterbottom2020datasetBias,arda2022}, despite the multi-modality of video datasets, not every class is audio-visually well-corresponded, \eg, Kinetics~\cite{kinetics} is visual modality biased.
Although several datasets are introduced as audio-visual datasets, many of the categories of these datasets may not be audio-visually correlated, such as ``civil defense siren'', ``wind noise'', or ``reversing beeps''. Moreover, even though categories are audio-visually correlated, they may not contain sufficiently dominant semantic signals that can properly bridge the sound-to-image generation. 
For example, ``people slurping'', ``people eating'', or ``people sneezing'' are similar in terms of containing human instances regardless of the category, while they completely differ in the audio modality. 
Such misalignment or weak correspondence in audio-visual modality may act like outliers and disturb the model learning to generate an image from the sound.
Thus, we conduct an analysis to identify which categories of audio-visual events are proper for the sound-to-image generation task.
\begin{figure}[tp]
    \centering
    \includegraphics[width=\linewidth]{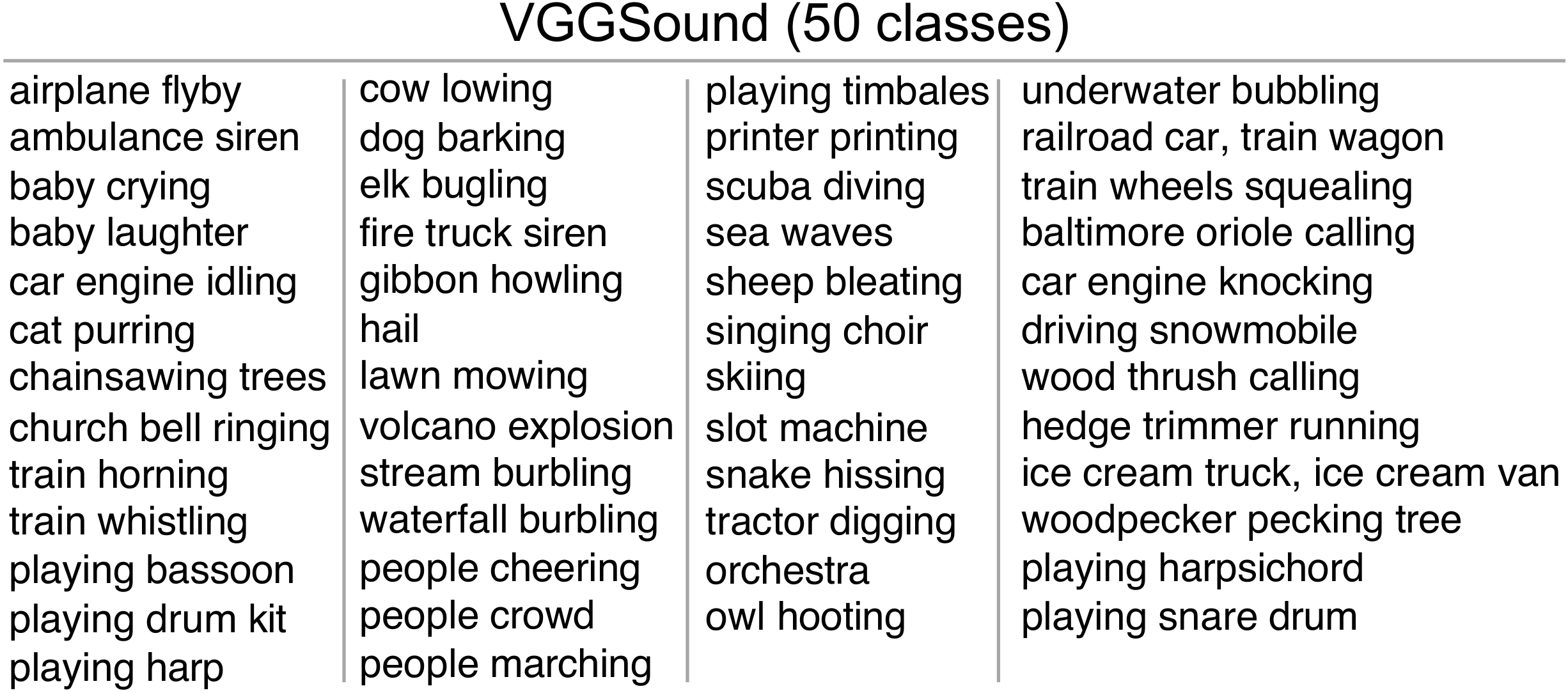}\vspace{-2mm}
    \caption{{\bf Selected audio-visual event categories.} We select 50 classes among VGGSound~\cite{vggsound} for sound-to-image generation.}
    \vspace{-3mm}
    \label{fig:class}
\end{figure}

We analyze the VGGSound~\cite{vggsound} dataset to find proper categories for the sound-to-image generation task, as large-scale benchmark datasets contain many in-the-wild videos and categories with very different characteristics.
For the analysis, we first train our model with all the categories in the VGGSound dataset. Then, we evaluate the $R@1$ of the generated images for each category using the CLIP~\cite{clip} retrieval metric introduced in the main paper.
The categories above a certain threshold in terms of $R@1$ performance naturally reveal plausible image generation quality. 

We discover that the categories related to action scenes are mostly excluded since our work focuses on a single frame generation task, which is more sensitive to the instance itself than the action in the scene.
In addition, we find that as we increase the number of categories, the image quality generated by our model degrades as it brings a high chance of including improper categories.

Given the analysis, we select the categories from VGGSound by sorting in $R@1$ performance and human perception.
We show the top-50 selected categories in \Fref{fig:class} that are suitable for the sound-to-image generation task. Since our analyses in the main paper are to see the quality of sound-to-image generation with much more diverse classes than the prior arts, we use those selected classes in all the experiments of the main paper. It contains more diverse categories with different levels of audio-visual correspondences compared to the existing methods that come with a small number of categories in which images and sounds are closely correlated.
Discovering more proper videos and filtering outliers to enhance the sound-to-image generation task is an interesting research direction and needs further investigation.

\begin{figure}[tp]
    \centering
    \footnotesize
    \includegraphics[width=\linewidth]{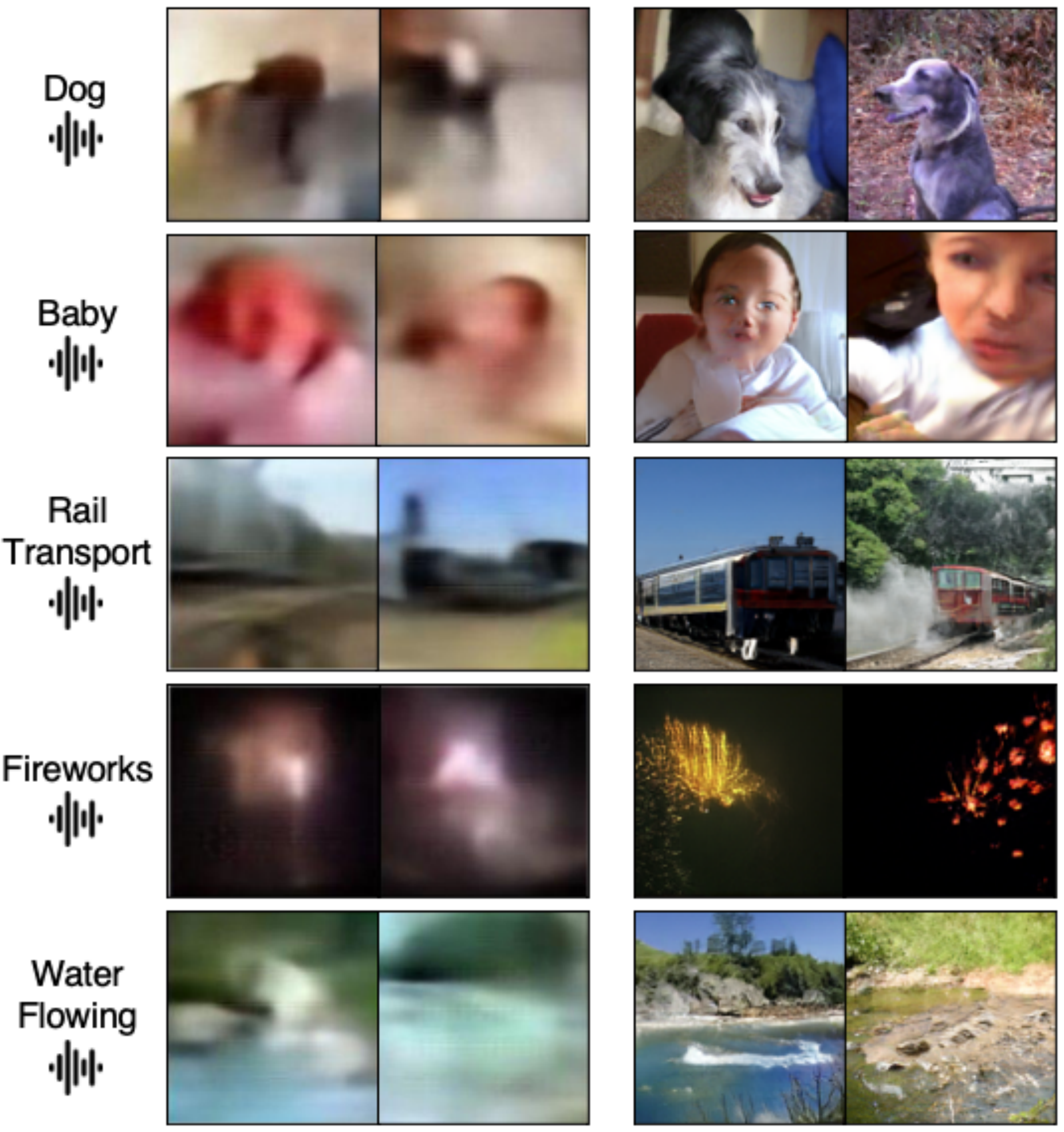}
    \blank{1.1cm} (a) S2I~\cite{s2i} \blank{2.3cm} (b) Ours\vspace{-2mm}
    \caption{\textbf{Qualitative comparison to S2I~\cite{s2i}.} We compare the generated images between (a) S2I and (b) Ours.}
    \vspace{-3mm}
    \label{fig:s2i}
\end{figure}

\begin{figure}[tp]
    \centering
    \footnotesize
    \includegraphics[width=\linewidth]{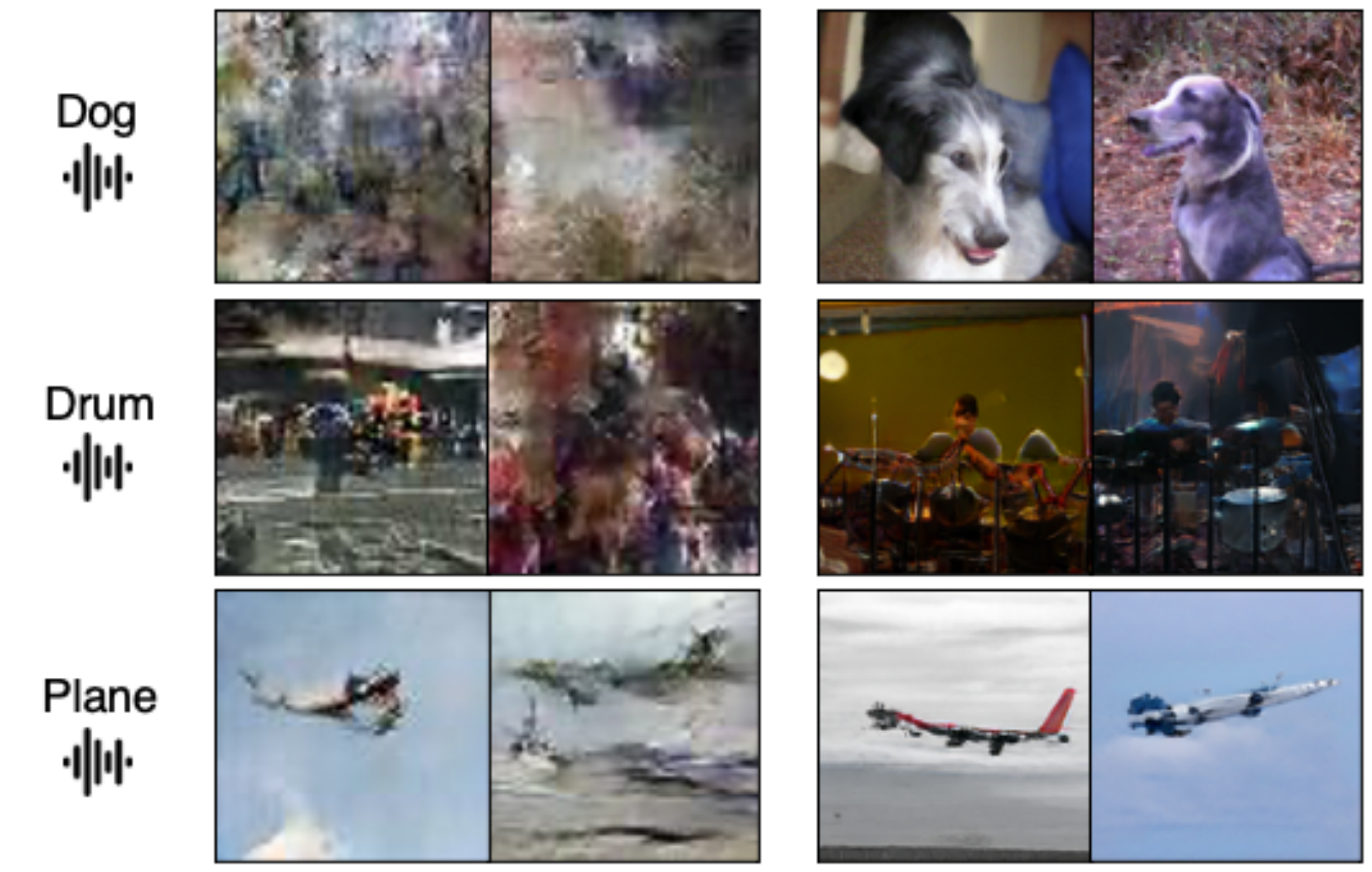}
    \blank{0.4cm} (a) Wan~\etal~\cite{towards_a2s} \blank{2cm} (b) Ours\vspace{-2mm}
    \caption{\textbf{Qualitative comparison to Wan~\etal~\cite{towards_a2s}.} We compare the generated images between (a) Wan~\etal and (b) Ours.}
    \vspace{-1mm}
    \label{fig:icassp}
\end{figure}

\section{Comparison with the Prior Arts}\label{sec:d}
We show qualitative comparison with our model and prior arts, Sound-to-Imagination (S2I)~\cite{s2i} and Wan~\etal\cite{towards_a2s}.
We obtain the generated images of the prior work directly from their published results.
Thus, the input audio for each generated image and the training dataset for each model is different. However, the purpose of this comparison is only to show how well our model and existing methods can generate images for given categories.
The image size varies depending on the models; our model generates $128\times128$, S2I generates $96\times96$, and Wan~\etal generate $64\times64$ pixels images. 

The comparison results of the overlapping sound categories for S2I and Wan~\etal are shown in Fig.\ref{fig:s2i}~and~\ref{fig:icassp}, respectively. 
While S2I preserves the coarse shapes of dogs or babies and contains scenes that are relevant to the input sound, the images are too blurry to clarify detailed depictions.
Wan~\etal also produces the coarse shape of the plane but fails to produce informative images on other categories. 
In contrast, our model consistently generates visually plausible and detailed images aligned with the given sound category. 
\begin{figure}[tp]
    \centering
    \includegraphics[width=\linewidth]{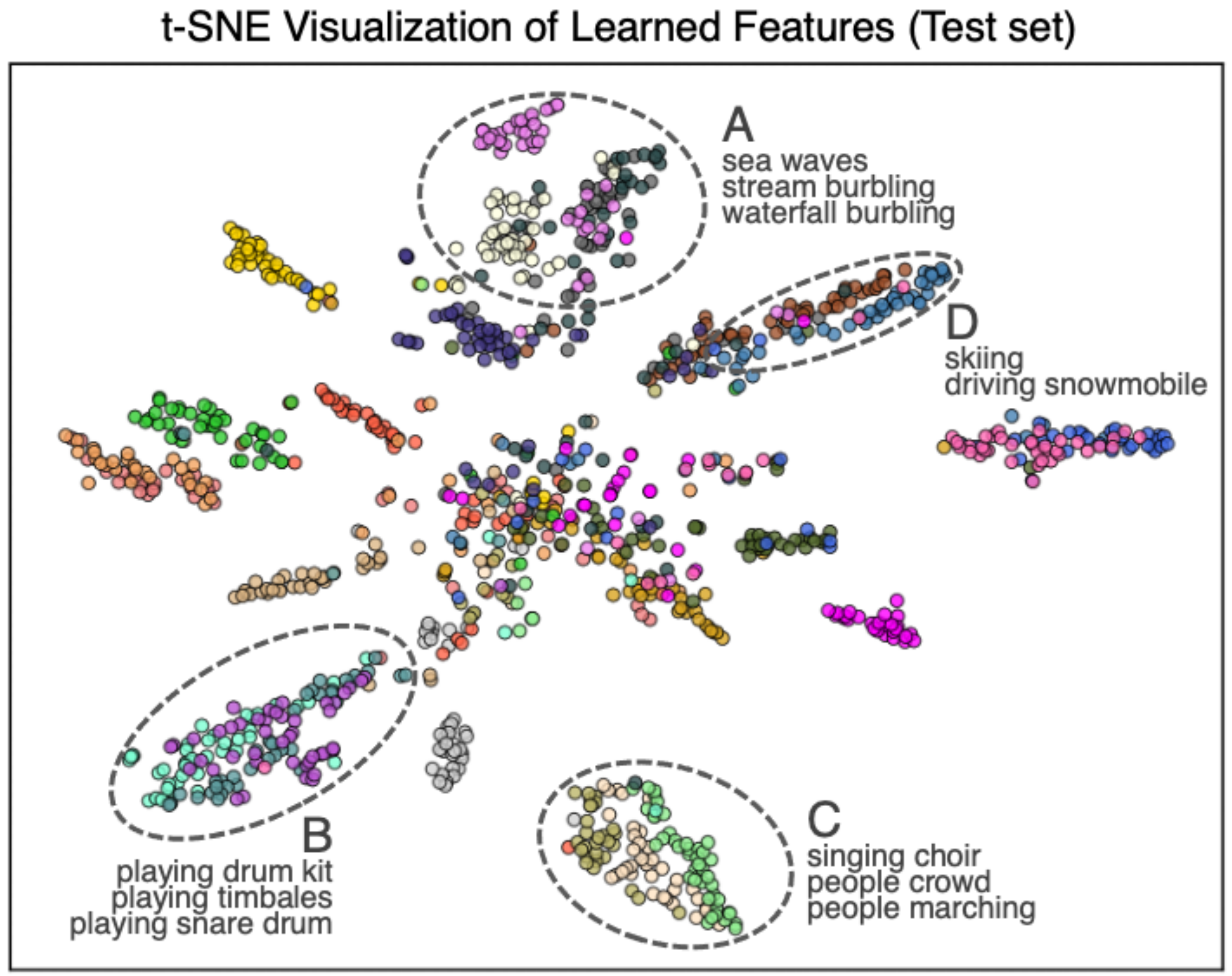}\vspace{-2mm}
    \caption{{\bf t-SNE visualization~\cite{tsne} of learned features.} We sample 25 classes from VGGSound and visualize the learned audio features of the test set. For visualization purposes only, we color the features in terms of class labels, and no labels are used for training.}
    \vspace{-2mm}
    \label{fig:tsne}
\end{figure}

\begin{figure*}[tp]
    \centering
    \includegraphics[width=\linewidth]{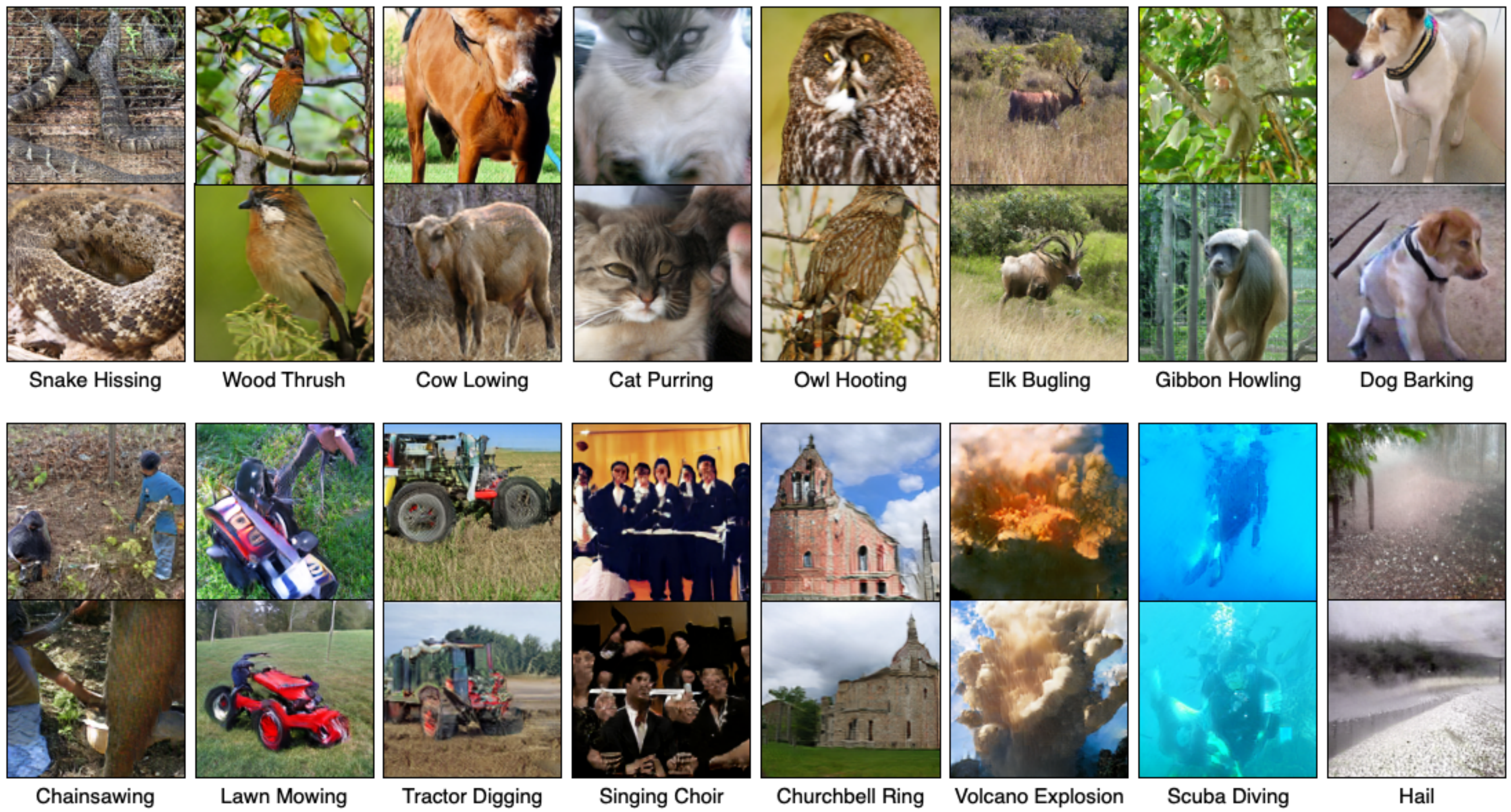}\vspace{-2mm}
        \caption{{\bf Qualitative results of sampled categories in VGGSound~\cite{vggsound} test set.} Sound2Scene can generate visually plausible images from diverse in-the-wild sounds. Note that our method does not use any class information during training and inference.}
    \label{fig:single_sup}
\end{figure*}

\begin{figure}[tp]
    \centering
    \includegraphics[width=0.96\linewidth]{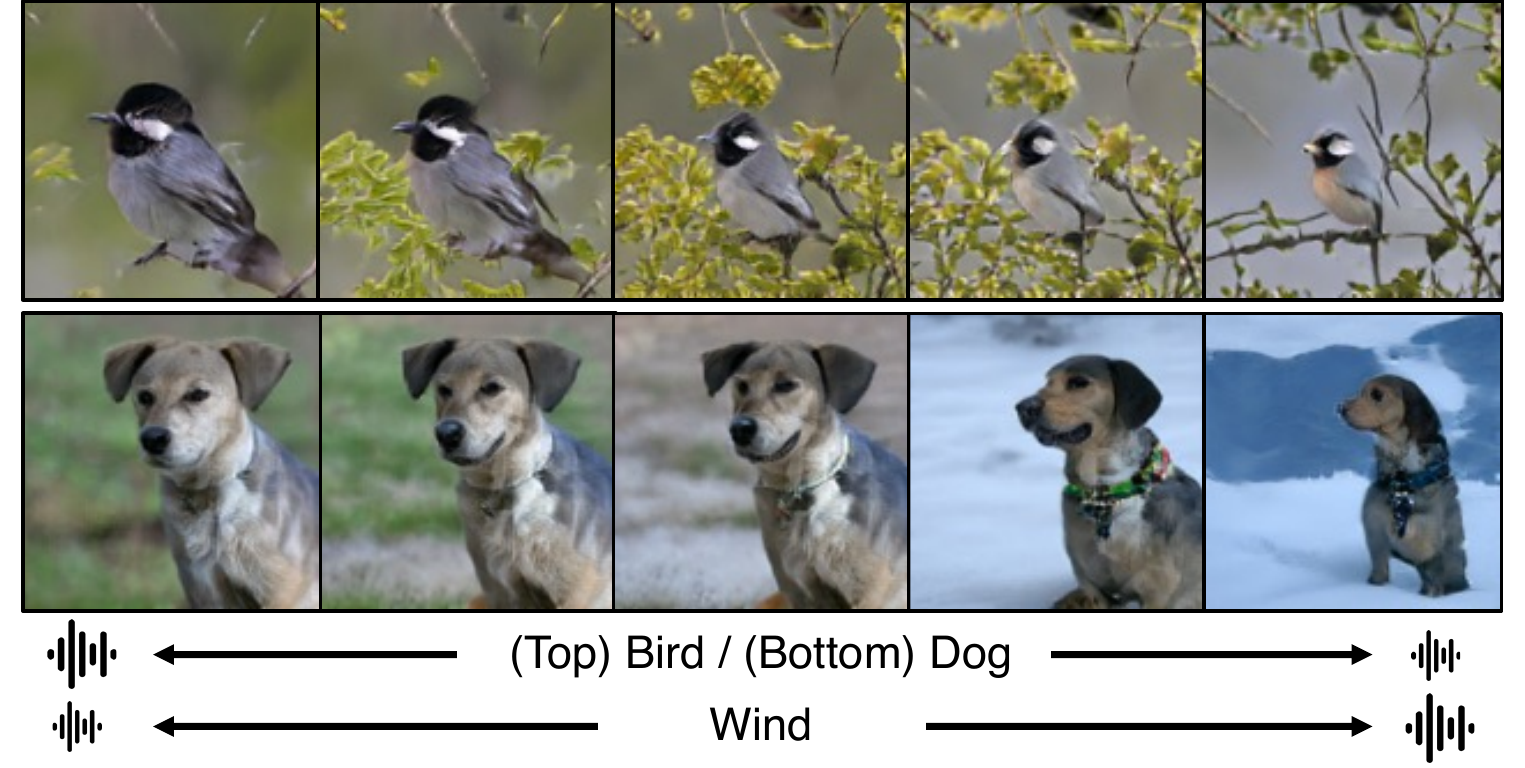}
    \vspace{-2mm}
    \caption{\textbf{Generated images by mixing multiple audios with volume changes in the \emph{waveform space}.}}
    \vspace{-5mm}
    \label{fig:volume_mix}
\end{figure}

\vspace{-2mm}
\section{Additional Qualitative Analysis and Results}\label{sec:e}

\paragraph{t-SNE visualization of audio features}
We show t-SNE visualization~\cite{tsne} of the learned features of our audio encoder in~\Fref{fig:tsne}.  
As shown, our audio encoder segregates input audios into clusters correlated with their audio-visually related classes. 
For example, three water-related clusters, which include similar visual scenes and also the sounds, are located closely (\textbf{A} in \Fref{fig:tsne}). In addition, drum-related videos are also similar in terms of audio-visual information and are located closely (\textbf{B} in \Fref{fig:tsne}).   
Although our model mostly embeds the input audios to audio-visually related clusters, several clusters are closely located in terms of visual information. This is expected as no class-level supervision is provided, but only the visual features are used.
For example, the sound of ``singing choir'', ``people crowd'', and ``people marching'' are different from each other and clustered separately, but they are closely located in terms of their visual similarity (\textbf{C} in \Fref{fig:tsne}), and the similar results are shown with ``skiing'' and ``driving snowmobile'' (\textbf{D} in \Fref{fig:tsne}).

\paragraph{Additional generated images from different sounds}
Additional qualitative results for generating images from single waveform are shown in~\Fref{fig:single_sup}. 
Each image is generated from different sounds without providing any class information to the model. 
As shown, our model can handle different categories of sounds, such as from animals, and vehicles, to diverse sceneries, and generate plausible results conditioned on the given sound. 
Generated images generally preserve the semantics of the scenes properly, such as the ``Chainsawing'' action appearing in the middle of the forest scene or ``Lawn Mowing'' images on grass instead of asphalt roads.

We further show the generated images by mixing multiple audios with volume changes in \Fref{fig:volume_mix}.
For example, by decreasing the volume of the ``Dog'' while increasing the ``Wind'' sound, a close-up shot of the dog starts disappearing and a wide-shot in the snowy environment (windy) with a smaller dog appears gradually.

\begin{figure}[tp]
    \centering
    \includegraphics[width=\linewidth]{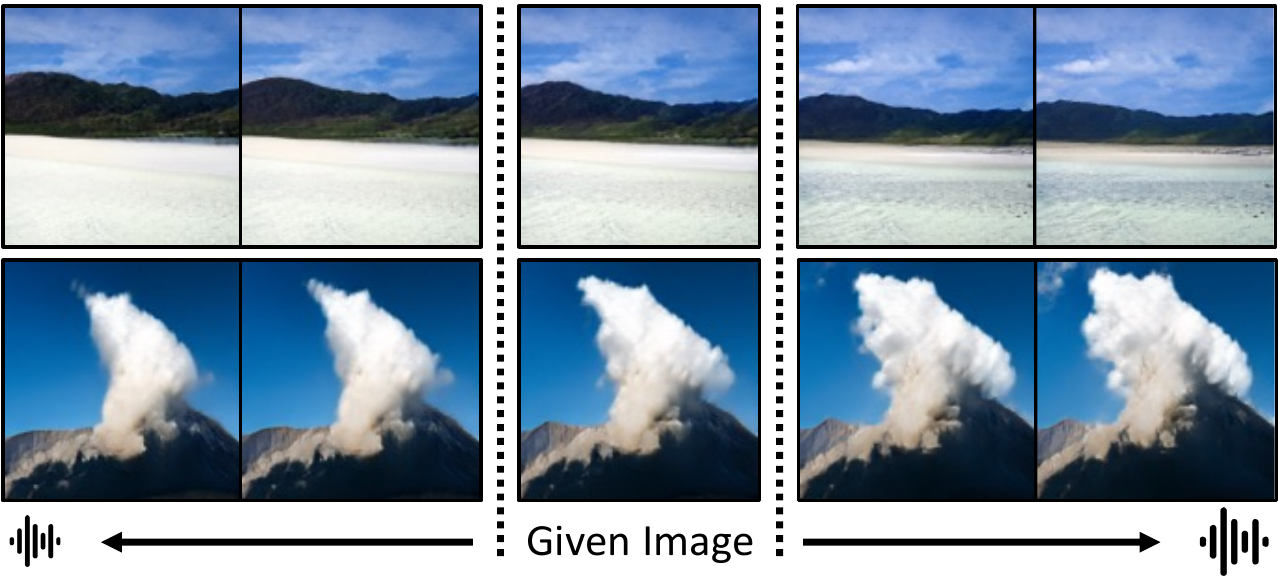}
    \caption{\textbf{Image editing by volume changes in the \emph{latent space}.} We move the extracted visual feature in the direction of the volume difference between the two audio features.}
    \vspace{-4mm}
    \label{fig:image_edit}
\end{figure}

\begin{figure*}[tp]
    \centering
    \includegraphics[width=\linewidth]{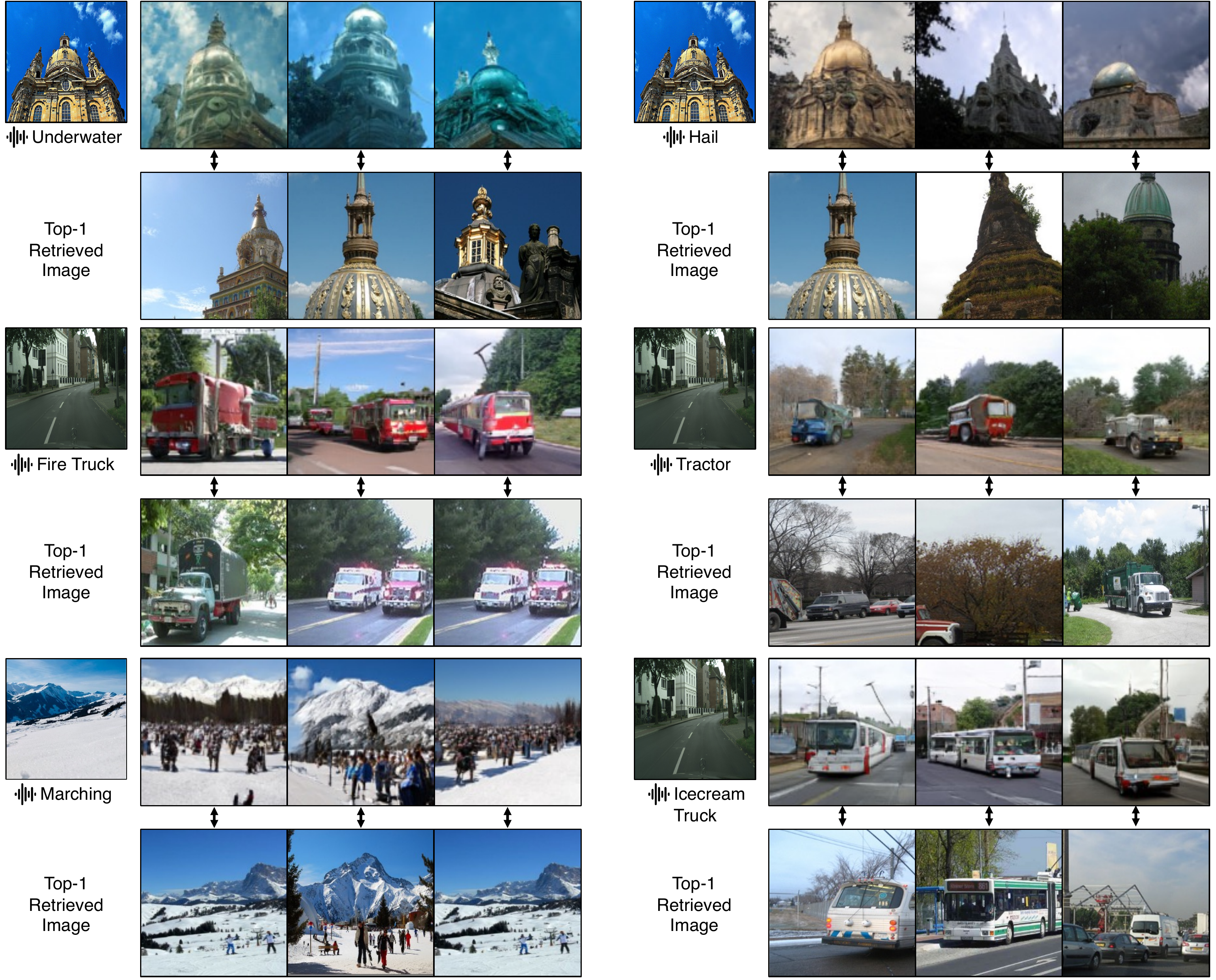}    
    \caption{{\bf Generated images conditioned on sound and image.} We simply interpolate between a visual feature and an audio feature in the joint latent embedding space. This interpolated feature is then fed to the image generator to generate a novel image. The first row per sample shows the generated images, while the row below contains the images from the ImageNet~\cite{imagenet} training set, which are retrieved by measuring cosine similarity in the latent space of the image encoder, $f_V(\cdot)$.}
    \label{fig:image_audio}
\end{figure*}

\paragraph{Additional generated images from latent manipulations}
As introduced in the main text, Sound2Scene provides latent space manipulations to generate images conditioned on both audio-visual signals.
For the first approach, we can edit the given image by the given paired audio.
We extract a visual feature and the noise vector by GAN inversion~\cite{gan_inv1, gan_inv2} and move the visual feature toward the volume change direction of the audio features.
Then, the manipulated visual feature and the noise vector are fed to the image generator, $G(\cdot)$, to generate an edited image.
As shown in \Fref{fig:image_edit}, the explosion of the given image gets smaller while we move the visual feature toward the volume-decreasing direction while getting bigger by moving toward the volume increase direction.

Furthermore, as a second approach, we can simply interpolate between the audio and visual features and generate a novel image by conditioning on both audio-visual signals, as shown in \Fref{fig:image_audio}.
We can stylize the building to be on a cloudy day or underwater, insert diverse vehicles on the road, or even insert people in the snowy fields.
Moreover, we compare the generated images with the closest samples in ImageNet~\cite{imagenet}. For example, we see that for generated images conditioned on the building and ``Underwater'' sound, no buildings in the closest images are underwater but with the blue sky; or we observe that the closest images rarely contain many people in the scene for the generated images conditioned on a snowy mountain and ``Marching'' sound. These examples show that our model generates new unique images rather than memorizing the training set.

\begin{figure*}[t]
\centering

\resizebox{1\linewidth}{!}{
\small
\begin{tabular}{cc}
\includegraphics[width=0.504\textwidth]{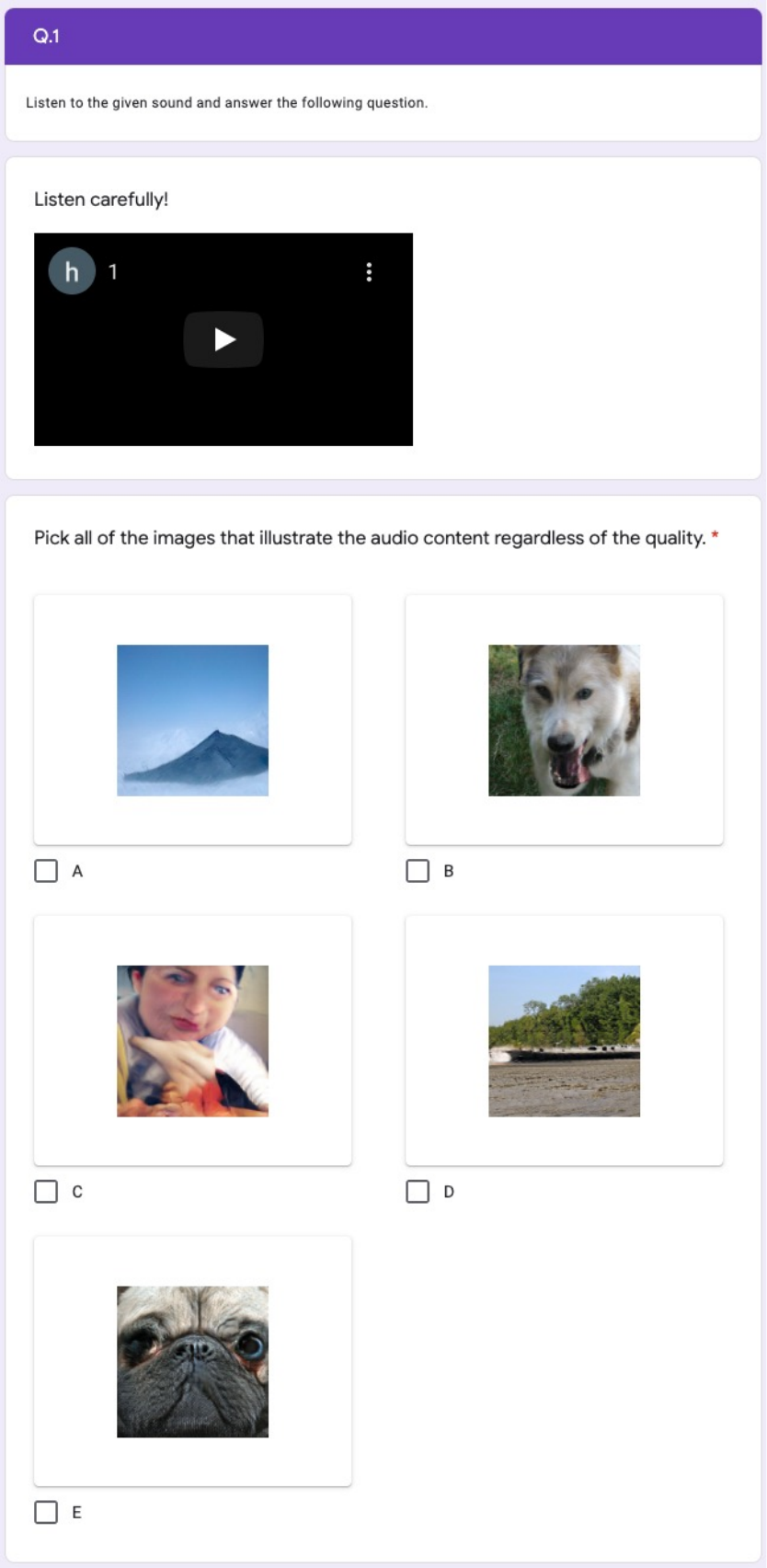}&\includegraphics[width=0.496\textwidth]{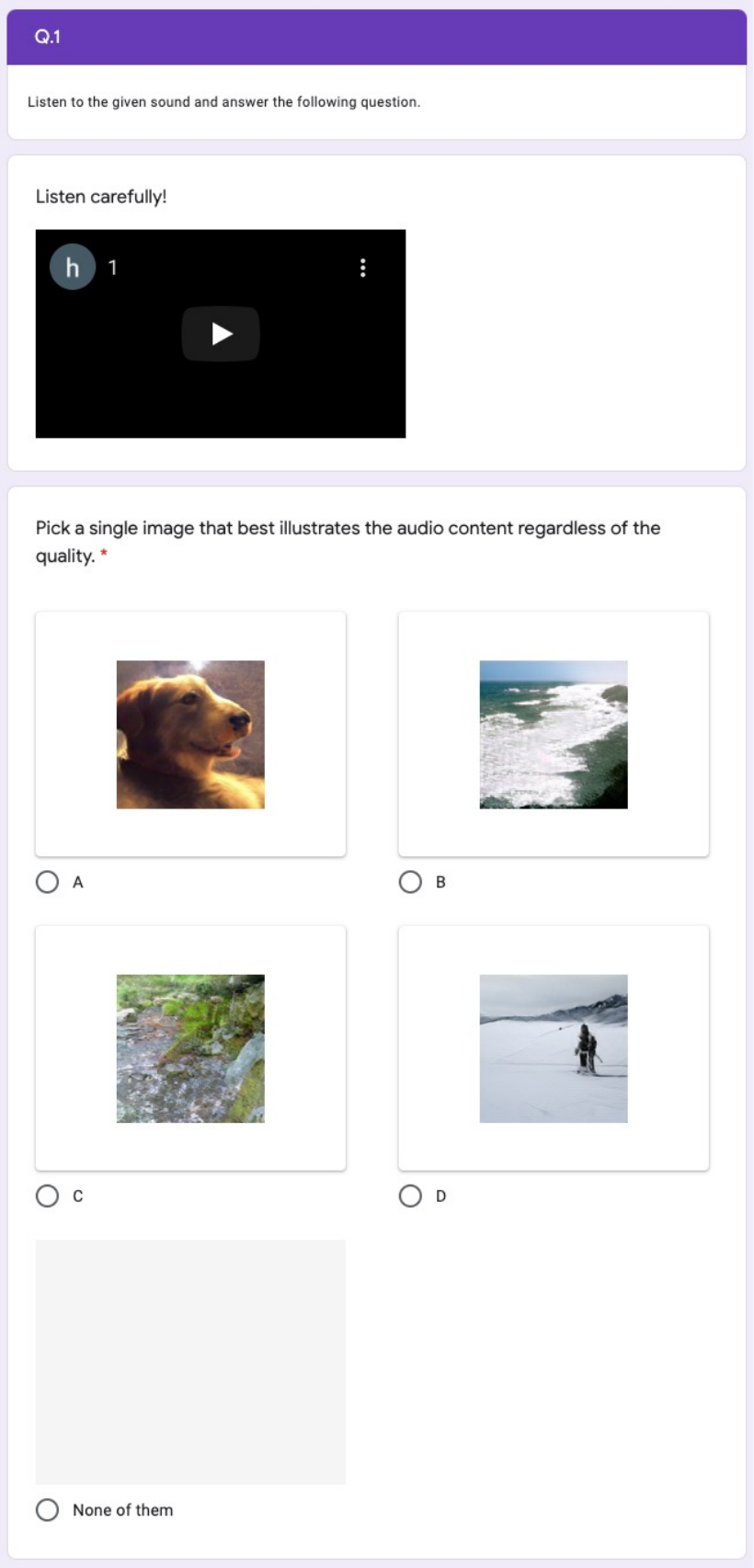}\\
(a) Comparison to ICGAN~\cite{icgan} &(b) Validation of proper image generation
\end{tabular}}
    \caption{
    \textbf{Examples of the user study.} We conduct the user study by comparing with ICGAN in (a) and validating the proper image generation in (b). The images provided in the user study are randomly ordered.
    }
    \label{figure:human_sup1}
\end{figure*}
\begin{figure*}[tp]
    \centering
    \small
    \includegraphics[width=\linewidth]{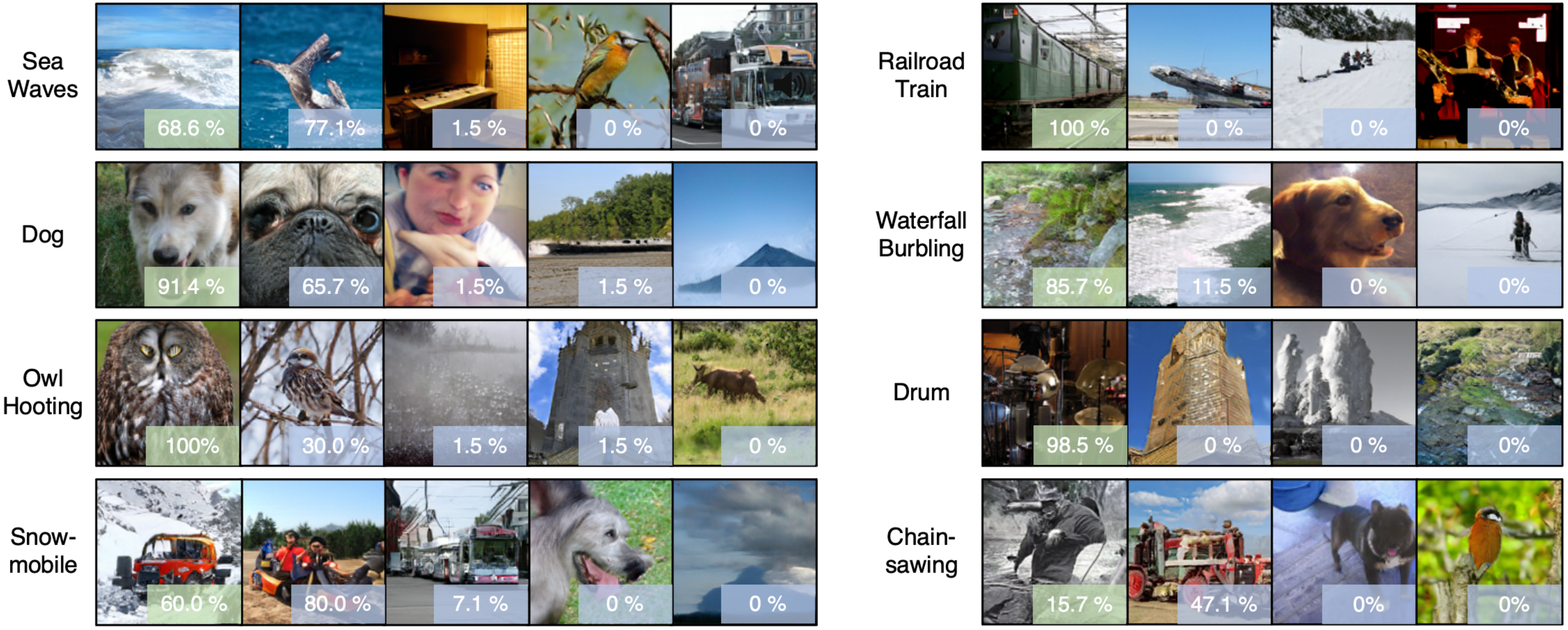}\vspace{-1.5mm}    
    \blank{0.3cm} Ours \blank{0.7cm} ICGAN \blank{1.25cm} Uncorrelated images \blank{3.1cm} Correct \blank{1.3cm} Uncorrelated images \\ \vspace{1.5mm}
    \blank{2.5cm} (a) Comparison to ICGAN~\cite{icgan} \blank{4.5cm} (b) Validation of proper image generation
    \caption{{\bf Samples of human evaluation results.} Each row denotes each question, and the corresponding audio is described with text. We compare the generated images with ICGAN in (a). Conditioned on the given audio-image pair (audio for ours and image for ICGAN), the two left images are generated by our model and ICGAN, respectively. Three remaining images are generated by our model but conditioned on uncorrelated category audios. Each percentage in the colored box states the recall probability of the generated image. We validate the proper image generation of our method in (b). All images are generated by our model, but only the first column is conditioned on the given sound. Each percentage in the colored box states the selection ratio of the participants.}
    \vspace{-2mm}
    \label{fig:human}
\end{figure*}

\section{Details of the User Study}\label{sec:f}
We conduct a user study to analyze the performance of our method from the human perception perspective. The user study questionnaire interface is shown in~\Fref{figure:human_sup1}. Users listen to the given audio, see the generated images and make a selection without any time limitation. This user study contains two experiments with 20 questions in each, as described in the main paper. The first experiment is about the comparison to ICGAN~\cite{icgan}. Audio and five images are given to the participants as \Fref{figure:human_sup1} (a). Among five images, two are generated by our model and ICGAN, respectively, from the given sound or its paired image. The rest are generated images from random categories of the sounds. The users are asked to pick all the images that illustrate the given sound. As shown in \Fref{fig:human}~(a), our generated images are more preferred to ICGAN. 
However, there are interesting results showing that the user study is highly subjective. For example, in the bottom row, even though only the image generated from our model can be considered as a snowmobile, users tend to pick the option that is more familiar than the snowmobile, a car-looking object, as the given audio is engine-like. 

In the second experiment, we validate how properly our model generates images for given audio.
Audio and four images are provided to the participants. Our model generates all four images, but only one image is from the given sound. Participants are asked to choose one image that best illustrates the given sound or check the \{None of them\} as in \Fref{figure:human_sup1} (b). The selection ratio in~\Fref{fig:human} (b) clearly shows that our model generates highly-correlated images to the given sounds from the human perspective.
We observe several interesting user subjectivities for making a selection. In the last row of \Fref{fig:human} (b), among four images, even though the generated image in the first column seems to be more related to the given sound (looks like a human is in the position of using a chainsaw), users select the second image containing a vehicle in the scene.
We assume that for people who are not experts in the audio-visual domain, it may be challenging to differentiate similar sounds (engine-like), \eg, chainsaw, tractor, and truck sounds. Nonetheless, the overall user studies support that our model generates visually plausible images corresponding to the given sound.


 \fi

\end{document}